\documentclass{article}
\usepackage{amsmath}

\usepackage{PRIMEarxiv}

\usepackage[utf8]{inputenc} 
\usepackage[T1]{fontenc}    
\usepackage{hyperref}       
\usepackage{url}            
\usepackage{booktabs}       
\usepackage{amsfonts}       
\usepackage{nicefrac}       
\usepackage{microtype}      
\usepackage{lipsum}
\usepackage{fancyhdr}       
\usepackage{graphicx}       
\graphicspath{{media/}}     
\usepackage{subcaption}
\usepackage{float} 
\usepackage{array}
\title{Probabilistic Surrogate Model for Accelerating the Design of Electric Vehicle Battery Enclosures for Crash Performance
}

\author{
  Shadab Anwar Shaikh \\
  Pacific Northwest National Laboratory \\
  Richland, WA\\
  \texttt{shadabanwar.shaikh@pnnl.gov}  \\
   \And
  Harish Cherukuri \\
  UNC Charlotte \\
  Charlotte, NC\\
  \texttt{hcheruku@charlotte.edu} 
    \And
    Kranthi Balusu \\
  Pacific Northwest National Laboratory \\
  Richland, WA\\
  \texttt{kranthi.balusu@pnnl.gov} \\
  \And
  Ram Devanathan \\
  Pacific Northwest National Laboratory\\
  Richland, WA\\
  \texttt{ram.devanathan@pnnl.gov} \\
  \And
  Ayoub Soulami \\
  Pacific Northwest National Laboratory\\
  Richland, WA\\
  \texttt{ayoub.soulami@pnnl.gov} \\
  }

\begin{document}
\maketitle

\begin{abstract}
This paper presents a probabilistic surrogate model for the accelerated design of electric vehicle battery enclosures with a focus on crash performance. The study integrates high-throughput finite element simulations and Gaussian Process Regression to develop a surrogate model that predicts crash parameters with high accuracy while providing uncertainty estimates. The model was trained using data generated from thermoforming and crash simulations over a range of material and process parameters. Validation against new simulation data demonstrated the model's predictive accuracy with mean absolute percentage errors within 8.08\% for all output variables. Additionally, a Monte Carlo uncertainty propagation study revealed the impact of input variability on  outputs. The results highlight the efficacy of the Gaussian Process Regression model in capturing complex relationships within the dataset, offering a robust and efficient tool for the design optimization of composite battery enclosures.
\end{abstract}


\section{Introduction}

In recent years, there has been a global push toward transitioning to electric vehicles (EVs) as a sustainable transportation alternative to gasoline-powered vehicles. Despite their environmental and operational benefits, broader adoption of EVs still is challenging. EVs are significantly heavier compared to their gasoline-powered counterparts. The weight of battery packs and their casings is a major factor that contributes to the increase in overall vehicle weight, which leads to reduced driving range \cite{iclodean2017a}. Furthermore, the structural integrity of the battery is critical, as damage to the battery pack can lead to thermal runaway, causing fires and explosions \cite{zhu2017LiBreview}.

To address these challenges, one innovative strategy could be the use of carbon fiber reinforced polymer (CFRP) composites\cite{azzopardi2023recent}. These materials are increasingly favored over traditional materials such as aluminum because of their superior strength-to-weight ratio, which is crucial for maintaining the structural integrity of battery enclosures while reducing the overall vehicle weight \cite{zhang2018aerspace, research2018a}. However, choosing CFRPs for battery enclosures introduces another set of challenges.

Design of CFRP enclosures requires careful adjustment of various material and manufacturing process parameters, which significantly affect performance of the final part \cite{kulkarni2023investigation} \cite{dorr2021simulation}. Choosing these parameters through trial-and-error experimentation is not feasible, as it can be both challenging and uneconomical. Fortunately, several physics-based simulation tools are available that can help alleviate these issues.

Traditional approaches to designing CFRP enclosures often depend on finite element analysis (FEA) tools such as PAM-FORM, which is known for its ability to accurately simulate the thermoforming process \cite{dorr2017benchmark}, and VPS, which provides precise modeling of complex damage behaviors that composites experience during impacts \cite{coppola2017a}. These tools can be integrated to enhance the analysis process. For example, PAM-FORM can be used to virtually manufacture CFRP battery enclosures through the thermoforming process, while VPS can be used simulate side pole crash impacts on enclosures. The combined use of VPS and PAMFORM allows seamless integration of thermoforming results into crash simulations, thus optimizing the development pipeline, as proposed by Kulkarni and Hale et. al \cite{kulkarni2023investigation}.

Despite the availability of these advanced simulation tools, identifying optimal design parameters for CFRP enclosures can be challenging because of the computational demands of these tools. To mitigate this issue, these tools can be coupled with data-driven approaches to enable speedy evaluation of potential combinations of design parameters to identify optimal settings.

In recent years, machine learning (ML) has made remarkable progress in tackling several types of problems \cite{he2016deep, devlin2018bert, shaikh2023recovering}. Deep learning algorithms have been adapted to handle tabular data, which typically present heterogeneous and non-sequential attributes, unlike image data that feature spatial hierarchies and local correlations that these models exploit effectively \cite{borisov2022deep}. As a result, tree-based models often outperform deep learning in handling tabular data scenarios \cite{grinsztajn2022tree, shwartz2022tabular}.

In addition to the strides made in ML for conventional data types, there has been notable progress in the deployment of these techniques within the domain of materials science \cite{gubernatis2018machine, morgan2020opportunities}. These advances facilitate faster exploration of design spaces, allowing designers to effectively find optimal design parameters. Particularly, Gaussian Processes (GPs) have proven to be an invaluable tool in this area due to their capability to model non-linear relationships and provide predictions that include estimates of uncertainty \cite{williams1995gaussian, wang2023intuitive, hoffer2022gaussian, tapia2018gaussian}. This feature makes GPs especially suitable for applications for which precise and reliable prediction is critical.

GPs have demonstrated their versatility by solving a diverse array of problems across manufacturing and material science domains. For example, Hoffer et al. \cite{hoffer2022gaussian} applied GPs to create a surrogate model for finite element method simulations used in the super-alloy forging process. Tapia et al. \cite{tapia2018gaussian} used Gaussian Process Regression (GPR) to determine  optimal process parameters in additive manufacturing of 316L steel with laser powder bed fusion. Zhou et al. \cite{zhou2007process} developed an adaptive surrogate model based on GPs for optimizing process parameters in injection molding simulations. Furthermore, Radaideh and Kozlowski \cite{radaideh2020surrogate} designed a GPR surrogate model for nuclear reactor simulations, which included uncertainty analysis. Saunders et al. \cite{saunders2021mechanical} constructed a functional GP surrogate to predict the mechanical behavior of additively manufactured microstructures. Noack et al. \cite{noack2020autonomous} proposed a GPR surrogate model equipped with anisotropic kernels and non-i.i.d. measurement noise for automating material discovery. Moreover, Chen et al. \cite{chen2021gaussian} employed GPR to model the constitutive relations of materials for stochastic structural analysis.

Numerous studies have been conducted to explore the application of computational techniques and ML in the design and optimization of composite materials \cite{liu2021machine, sharma2022advances}. Zhang et al. \cite{zhang2022mechanical} introduced a hybrid approach combining FEA and ML, specifically using neural networks and random forest, to predict the mechanical properties of composite laminates. Balcioglu et al. \cite{balciouglu2021comparison} investigated the fracture mechanics of laminated composites, employing experimental methods, FEA, and ML algorithms to focus on Mode I, Mode I/II, and Mode II loading situations. The goal of their research was to assess the comparative effectiveness of FEA and ML against traditional experimental approaches in analyzing the fracture behavior of composites. Additionally, Milad et al. \cite{milad2022development} evaluated the performance of three ensemble ML models in predicting the strain enhancement ratio of fiber-reinforced polymer composites, drawing on a comprehensive dataset of 729 experiments from existing literature.

Although FEA and ML have been widely applied in the design and manufacturing of composites, their potential to accelerate the design of composite battery enclosures specifically for crashworthiness has not been extensively investigated. Addressing this research gap, the initial phase of this study was focused on creating a dataset via high-throughput simulations that are both time and resource demanding, adhering to the simulation workflow established by Kulkarni and Hale et al. \cite{kulkarni2023investigation}. Subsequently, tree-based ensemble methods were developed, which were trained on this newly generated dataset \cite{shaikh2023finite}. This approach aimed to leverage the predictive power of tree-based models to optimize the crash performance of composite battery enclosures.

This study is dedicated to the development of a probabilistic surrogate model using GPs that not only delivers predictions but also provides estimates of associated uncertainties. The surrogate model is designed to facilitate and accelerate design space exploration, offering a viable path for quickly optimizing the design of composite structures to meet specific performance criteria. By integrating uncertainty estimates, this model enhances decision-making processes, enabling more informed and reliable design choices in the development of composite materials and structures.

\section{Background}

In this section, we provide an overview of the current simulation workflow used in the design of battery enclosures, highlighting the time-intensive nature of the process. Additionally, we describe the mathematical underpinnings of GPR and surrogate modeling.

\subsection{Current simulation workflow for enclosure design}
\begin{figure}[h]
\centering
\includegraphics[width = 1\linewidth]{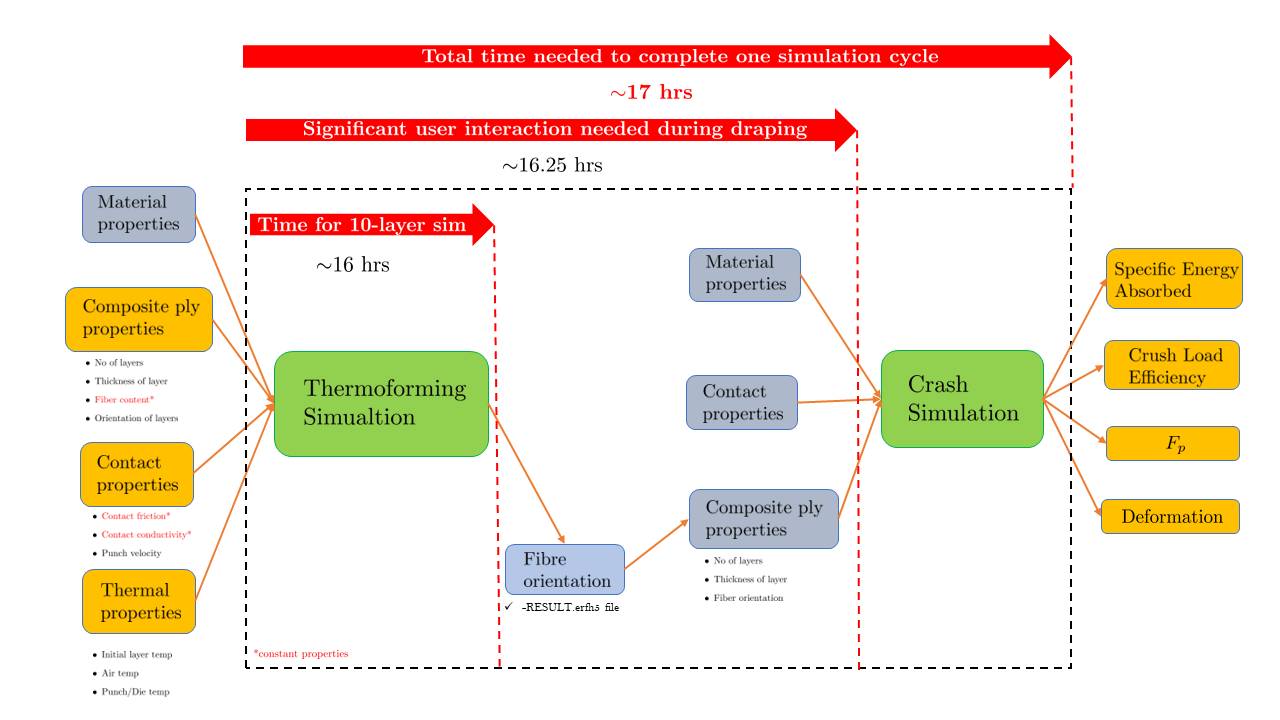}
\caption{Schematic of the current simulation workflow for enclosure design \cite{shaikh2023finite}}
\label{fig:simflow}
\end{figure}

The current simulation workflow, depicted in Figure \ref{fig:simflow}, begins with a thermoforming simulation to virtually manufacture the battery enclosures. This is followed by a crash simulation that simulates a side-pole impact test on the formed part, and finally, the extraction of crucial crash parameters.

However, the main challenge with this workflow is the time required to complete each step. For instance, the thermoforming simulation of a 10-layer composite laminate typically takes approximately 16 hours on a 16-core high-performance computing (HPC) unit. Subsequently, the crash simulation requires about 45 minutes on an 8-core 11\textsuperscript{th} Gen Intel (R) i7@3GHz PC. Moreover, linking these two simulations necessitates manually transferring the results from the thermoforming to the crash simulation, a process that demands significant user interaction. We found that completing one full simulation run takes around 17 hours, requiring substantial user involvement throughout. This extensive time commitment complicates the decision-making process, underscoring the need for a surrogate model to significantly reduce the overall simulation time.

\subsection{Surrogate Modeling using ML}

As discussed in \cite{shadabthesis2024}, a computational model can be conceptualized from a mathematical standpoint as a 'black-box' function, denoted as $\mathcal{M}$. This function maps the input parameters $\mathbf{X} \in \mathbb{R}^{n \times d_1}$ of the system to the outputs $\mathbf{Y} \in \mathbb{R}^{n \times d_2}$, represented by the equation:
\begin{equation}
\mathbf{Y} = \mathcal{M} (\mathbf{X})
\end{equation}
where $\mathcal{M}: \mathbf{X} \mapsto \mathbf{Y}$, with $d_1$ and $d_2$ being the dimensions of the input and output spaces, respectively.

In this study, $\mathcal{M}$ abstractly represents a finite element simulation chain for designing enclosures for crash performance. The model inputs include material parameters such as the number of composite layers $n_{ls}$, thickness of each layer $t_l$, fiber orientation $\mathbf{\Phi}_{fib}$, and process parameters for the thermoforming process, including the punch velocity $v_p$, initial temperatures of the composite layer $T_i$, punch/die $T_{pd}$, and air $T_{air}$.

Typically, the finite element model predicts outputs such as specific energy absorption ($\mathrm{SEA}$), crush load efficiency ($\mathrm{CLE}$), peak force ($F_p$), and the intrusion of the side pole inside the battery enclosure after impact ($\mathrm{\Delta Y_{node}}$).

Conventionally in surrogate modeling, the objective is to develop an approximate function, or a simulator $\mathcal{\hat{M}}$, that mimics the behavior of the actual function $\mathcal{M}$. This process is succinctly expressed by the equation:
\begin{equation} \label{eq:ml_surrogate}
\mathbf{\hat{Y}} = \mathcal{\hat{M}} (\mathbf{X}; \mathbf{\Theta})
\end{equation}
where $\mathbf{\Theta}$ represents the collection of model parameters. The primary objective is to find optimal $\mathbf{\Theta}$ by utilizing the dataset $\mathcal{D}$ and optimization strategies such that $\hat{\mathbf{Y}} \approx \mathbf{Y}$. The model should not only closely fit the given data but also capture the underlying patterns and relationships inherent in the data, enabling it to generalize well to unseen data.

The dataset is constructed by executing the actual model $\mathcal{M}$ on a collection of input parameters $\mathbf{X} = \left[ \mathbf{x}^{(1)}, \mathbf{x}^{(2)}, \ldots, \mathbf{x}^{(N)} \right]^{\mathrm{T}} \in \mathbb{R}^{N \times 10}$, where $\mathbf{x} = \left[n_{ls}, t_l, v_p, T_i, T_{pd}, T_{air}, \mathbf{\Phi}_{fib} \right]$, with $\mathbf{\Phi}_{fib}$ being a vector of size 4, for a finite number of runs $N$. Subsequently, the predicted output $\mathbf{Y}$, where $\mathbf{Y} = \left[\mathbf{y}^{(1)}, \mathbf{y}^{(2)}, \ldots, \mathbf{y}^{(N)}\right]^{\mathrm{T}} \in \mathbb{R}^{N \times 4}$, where $\mathbf{y} = \left[\mathrm{SEA}, \mathrm{CLE}, F_p, \mathrm{\Delta Y_{node}} \right]$ is aggregated to assemble the dataset $\mathcal{D} = \{\mathbf{X}, \mathbf{Y}\}$.

Furthermore, the surrogate model $\mathrm{\hat{\mathcal{M}}}$ can represent any ML algorithm, including but not limited to linear regression, neural networks, support vector machines and GPR. In the context of this study, our approach begins with the application of linear regression, progressing subsequently to Gaussian processes due to their inherent capacity to effectively capture non-linear patterns within the dataset.

\subsection{Gaussian Process Regression}\label{sec:gpr}

Based on the discussion in \cite{nemani2023uncertainty}, a Gaussian process is a random process $f(\mathbf{x})$ where $\mathbf{x} \in \mathbb{R}^d$, with $d$ being a dimension of the input space, such that any finite subset of these inputs points given by $\mathbf{X} =\left[\mathbf{x}^{(1)}, \ldots, \mathbf{x}^{(n)}\right]^{\mathrm{T}} \in \mathbb{R}^{n \times d}$, and their corresponding output $f\left(\mathbf{X} \right)=\left[f\left(\mathbf{x}^{(1)}\right), \ldots, f\left(\mathbf{x}^{(n)}\right)\right]^{\mathrm{T}} \in \mathbb{R}^n$ is also a joint Gaussian distribution.

Gaussian process regression starts with a prior of the unknown function given by $f(\mathbf{x}) \sim \mathcal{G P}\left(m(\mathbf{x}), k\left(\mathbf{x}, \mathbf{x}^{\prime}\right)\right)$, which is fully characterized by a mean function $m(\mathbf{x}): \mathbb{R}^d \mapsto \mathbb{R}$, that defines the prior mean of $f$ for certain input points $\mathbf{x}$, given by; \begin{equation}
    m(\mathbf{x})=\mathbb{E}[f(\mathbf{x})]
\end{equation} and covariance function $k\left(\mathbf{x}, \mathbf{x}^{\prime}\right): \mathbb{R}^d \times \mathbb{R}^d \mapsto \mathbb{R}$, which also is known as a kernel of GPR, that basically captures the linear dependence of function values at two input points $\mathbf{x}$ and $\mathbf{x}^{\prime}$ and given by \begin{equation}
k\left(\mathbf{x}, \mathbf{x}^{\prime}\right)=\mathbb{E}\left[(f(\mathbf{x})-m(\mathbf{x}))\left(f\left(\mathbf{x}^{\prime}\right)-m\left(\mathbf{x}^{\prime}\right)\right)\right].
\end{equation} To make posterior calculations manageable, the prior mean $m(\mathbf{x})$ is often set to zero at all points i.e. $m(\mathbf{x}) = 0$. In this case, the covariance function defines the shape of the function sampled from the prior.

The most commonly used kernel (i.e., covariance function) is the radial basis function kernel (a.k.a., the squared exponential kernel) given by:\begin{equation}\label{eq:sekernel}
k\left(\mathbf{x}, \mathbf{x}^{\prime}\right)=\sigma_f^2 \exp \left(-\frac{\left\|\mathbf{x}-\mathbf{x}^{\prime}\right\|^2}{2 l^2}\right)
\end{equation} where $\sigma_f$ and $l$ are the are two hyper-parameters of kernel (or more specifically the GPR model) known as the signal variance and the length-scale, respectively. Additionally, this is a special case of general class of kernels called Matérn kernels that are given by:  \begin{equation}
k\left(\mathbf{x}, \mathbf{x}^{\prime}\right)=\sigma_{f}^2 \frac{2^{1-\nu}}{\Gamma(\nu)}\left(\frac{\sqrt{2 \nu}\left\|\mathbf{x}-\mathbf{x}^{\prime}\right\|}{l}\right)^\nu K_\nu\left(\frac{\sqrt{2 \nu}\left\|\mathbf{x}-\mathbf{x}^{\prime}\right\|}{l}\right)
\end{equation} where $\nu$ controls the smoothness, $\Gamma$ is the gamma function, and $K_\nu$ is a modified Bessel function. As $\nu \to \infty$, this become equivalent to the radial basis kernel. 

The kernel in Eq. \ref{eq:sekernel} uses same length-scale across all input dimensions $d$. Another approach involves using a different length-scale $l_i$ for each input dimension $x_i$, known as automatic relevance determination (ARD), given by: \begin{equation}
    k\left(\mathbf{x}, \mathbf{x}^{\prime}\right)=\sigma_f^2 \exp \left(-\frac{1}{2} \sum_{i=1}^d \frac{\left(x_i-x_i^{\prime}\right)^2}{l_i^2}\right)
\end{equation} where the kernel parameters are the length scales $l_1, \ldots, l_d$ and the signal amplitude, $\sigma_f$. The value of the length-scale determines the relevance of each input feature to the GPR model. This kernel is also known as an-isotropic variant of the squared exponential kernel. 

In GPR, the task is to infer relationship between the inputs $\mathbf{X}$ and target $\mathbf{y}$ (i.e. $\mathbf{y} =\left[y^{(1)}, \ldots, y^{(n)}\right]^{\mathrm{T}} \in \mathbb{R}^{n}$) where $n$ is number of noisy observations, assembled as a dataset given by $\mathcal{D}=\left\{\mathbf{X}, \mathbf{y}\right\}$. Following the definition of Gaussian process, the value of functions at the training inputs $\mathbf{X}$ and new unseen input points $\mathbf{X}^{*}$ i.e. $\mathbf{f}^{*}$ are jointly Gaussian, written as 

\begin{equation}
\left[\begin{array}{c}
\mathbf{y} \\
\mathbf{f}^{*}
\end{array}\right] \sim \mathcal{N}\left(\mathbf{0},\left[\begin{array}{cc}
\mathbf{K}_{\mathbf{X}, \mathbf{X}}+\sigma_{n}^2 \mathbf{I} & \mathbf{K}_{\mathbf{X}, \mathbf{X}^*} \\
\mathbf{K}_{\mathbf{X}^*, \mathbf{X}} & \mathbf{K}_{\mathbf{X}^*, \mathbf{X}^*}
\end{array}\right]\right) .
\end{equation} where $\mathbf{K}_{\mathbf{X}, \mathbf{X}}$ is the covariance matrix between the training points, $\mathbf{K}_{\mathbf{X}, \mathbf{X}^{*}}$ and $\mathbf{K}_{ \mathbf{X}^{*}, \mathbf{X}}$ are the covariance matrices between the training points and new points (also known as cross-covariance matrix), $\mathbf{K}_{ \mathbf{X}^{*}, \mathbf{X}^{*}}$ is the covariance matrix between the new points, and $\sigma_{n}^2$ represents variance of an additive independent and identically distributed (i.i.d.) Gaussian noise with zero mean.

The predictions can be made by drawing the samples from the posterior distribution, which is also multivariate Gaussian by definition \cite{williams1995gaussian}, with posterior mean given by \begin{equation}\label{eq:postmean}
\mathrm{mean}({\mathbf{f}}^*)=\mathbf{K}_{\mathbf{X}, \mathbf{X}^*}^{\mathrm{T}}\left(\mathbf{K}_{\mathbf{X}, \mathbf{X}}+\sigma_{n}^2 \mathbf{I}\right)^{-1} \mathbf{y},
\end{equation} and covariance \begin{equation}\label{eq:postcov}
\mathrm{cov} \left(\mathbf{f}^* \right)=\mathbf{K}_{\mathbf{X}^*, \mathbf{X}^*}-\mathbf{K}_{\mathbf{X}, \mathbf{X}^*}^{\mathrm{T}}\left(\mathbf{K}_{\mathbf{X}, \mathbf{X}}+\sigma_{n}^2 \mathbf{I}\right)^{-1} \mathbf{K}_{\mathbf{X}, \mathbf{X}^*}.
\end{equation} An important point to note here is calculating mean and covariance using equations \ref{eq:postmean} and \ref{eq:postcov} involve inverting the matrix which may be often large and ill-conditioned. Further, to alleviate these issues oftentimes a small regularization term is on the diagonal elements of the matrix. In addition, $\sigma_{n}^2$ in the equations also serves as a regularization term.

In a GPR model, training implies estimating unknown hyper-parameters (i.e. optimal model parameters) $\Theta = [l, \sigma_f, \sigma_n ]^{\mathrm{T}}$, where $l$, $\sigma_f$, and $\sigma_n$ are length scale, signal amplitude and noise variance respectively, typically involve maximizing log marginal likelihood given by : \begin{equation}
\log p\left(\mathbf{y} \mid \mathbf{X}, \boldsymbol{\Theta}\right)=-\frac{1}{2}(\mathbf{y}^{\mathrm{T}}\left(\mathbf{K}_{\mathbf{X}, \mathbf{X}}+\sigma_{n}^2 \mathbf{I}\right)^{-1} \mathbf{y})-\frac{1}{2} \log \left|\mathbf{K}_{\mathbf{X}, \mathbf{X}}+\sigma_{n}^2 \mathbf{I}\right| -\frac{N}{2} \log (2 \pi).
\end{equation} In this equation, the first term quantifies how well the model fits the data and the second term quantifies model complexity and it is added to prefer simpler over complex during training.

\section{Methodology}\label{sec:headings}

This section highlights the methodology  we employed to develop the surrogate model. Initially, we conducted high-throughput finite element simulations to create a comprehensive dataset $\mathcal{D}$, consisting of material and process parameters $\mathbf{X}$ and corresponding crash outputs $\mathbf{Y}$. Subsequently, the surrogate model $\mathcal{\hat{M}}$ was constructed by fitting an ML model to this dataset and identifying the optimal model parameters $\mathbf{\Theta}$.

\begin{figure}[h]
    \centering
    \includegraphics[width = 1\linewidth]{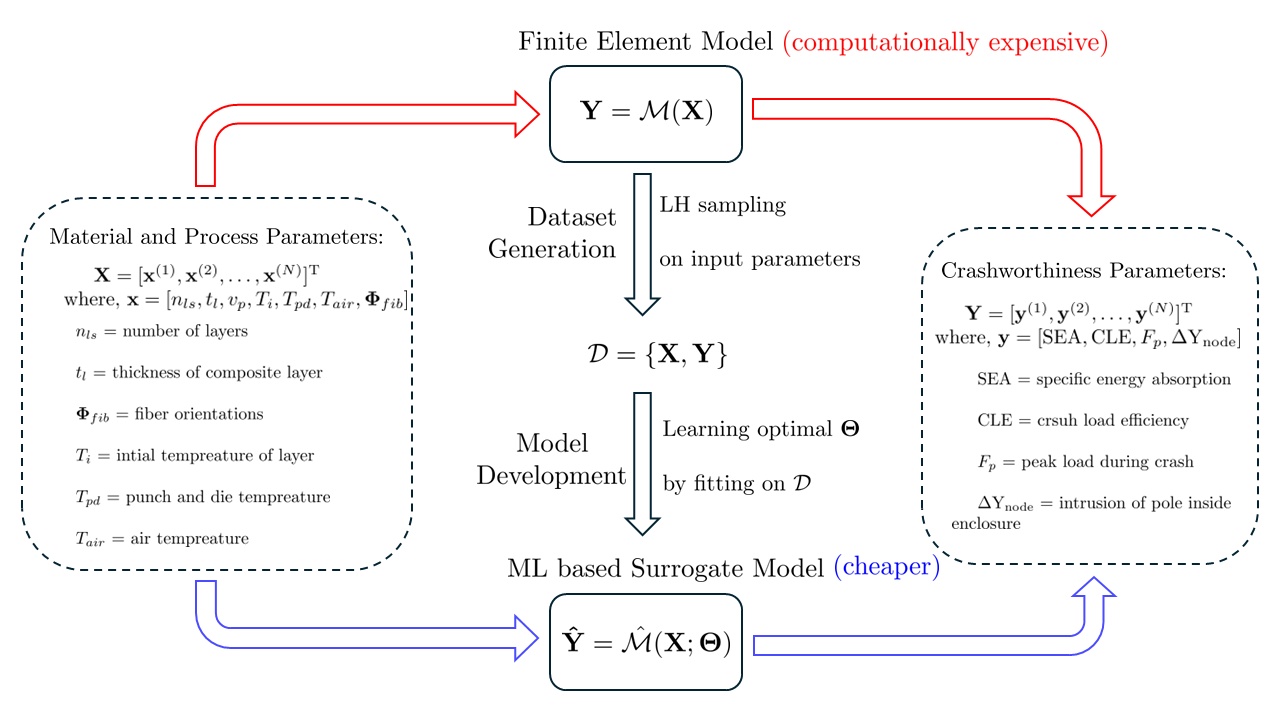}
    \caption{Summary of overall methodology}
    \label{fig:summary_methodology}
\end{figure}
Figure \ref{fig:summary_methodology} illustrates the overall methodology used in this study. Each step is discussed in detail in the following subsections.

\subsection{Data generation for surrogate model}

\subsubsection{Geometry of the battery enclosure}

\begin{figure}[h]
\centering
\subfloat[Enclosure]{\includegraphics[width=0.35\textwidth]{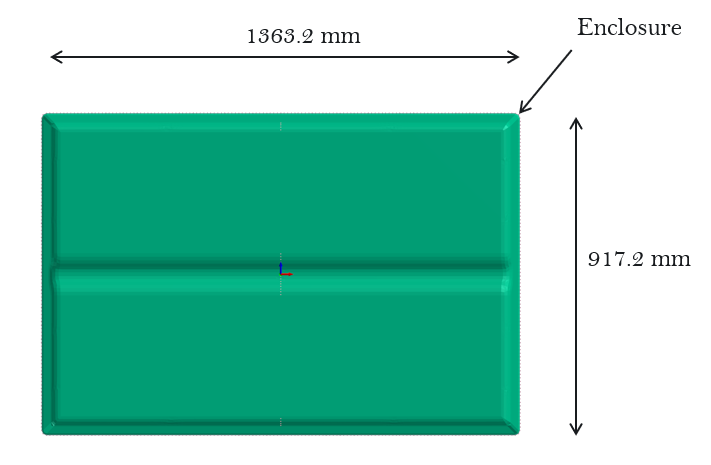}}
\subfloat[with ribs]{\includegraphics[width=0.35\textwidth]{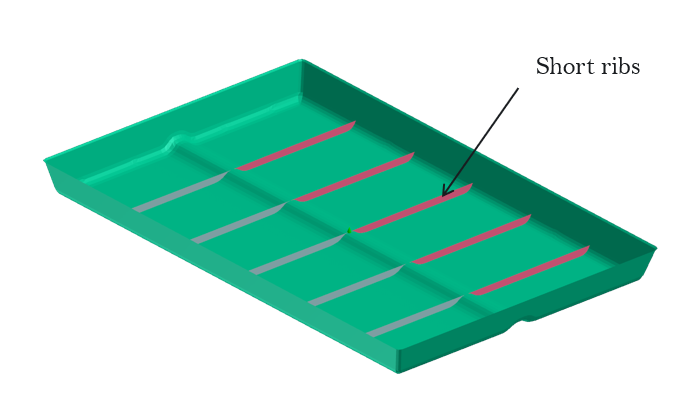}}\
\subfloat[with lid]{\includegraphics[width=0.35\textwidth]{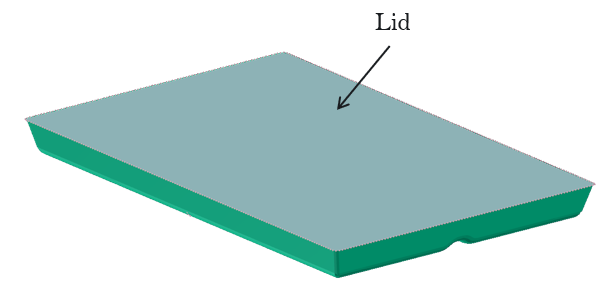}}
\caption{Geometry of the battery enclosure used in the analysis \cite{kulkarni2023investigation}}
\label{fig:batgeo}
\end{figure}

The geometry of the battery enclosure, as shown in Figure \ref{fig:batgeo}, was used in  subsequent simulation setups. We based the dimensions of the enclosure on those of an actual battery module \cite{cui2020application, lu2016muti, kulkarni2023investigation}; however, to make the analysis computationally more manageable, we incorporated several simplifications .

First, no internal or external mountings were included in the model. Additionally, a nonstructural mass of approximately 100 kg was added to simulate the realistic weight of a battery pack. For added structural support, the enclosure was internally reinforced with ribs, as depicted in Figure \ref{fig:batgeo}(b). Furthermore, a 15-degree relief angle was introduced at the ends of the enclosure to facilitate the removal of the punch during the thermoforming manufacturing process. The entire assembly was covered with a lid, as shown in Figure \ref{fig:batgeo}(c). Both the lids and ribs were assumed to be made of elasto-plastic material.

\subsubsection{Thermoforming simulation setup}\label{sec:thermo}

The geometry of the die was based on the overall dimensions of the enclosure. Additionally, to address convergence issues caused by excessive wrinkling of the composite sheet during the forming stroke, a custom outline of the ply with a cutout relief was created. The dimensions of the cutout were determined using the following formula:
\begin{equation}
c^2 = \frac{S_p - A_s}{4},
\end{equation}
where $c$ is the length of the cutout, $S_p$ is the outer surface area of the punch, and $A_s$ is the area of the composite sheet. We found that a square cutout measuring 105 x 105 mm for a sheet size of 1,500 x 1,052 mm resolved all convergence issues.

\begin{figure}[h]
\centering
\subfloat[]{\includegraphics[width = 0.4 \linewidth]{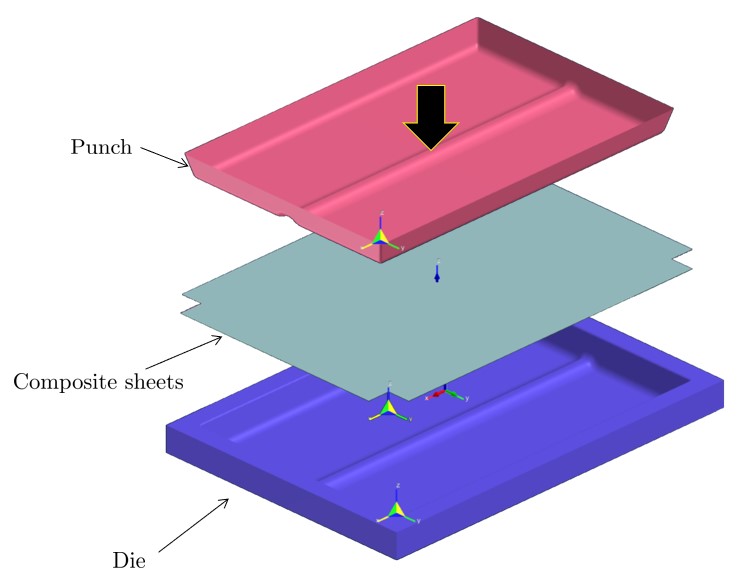}}
\subfloat[]{\includegraphics[width = 0.6 \linewidth]{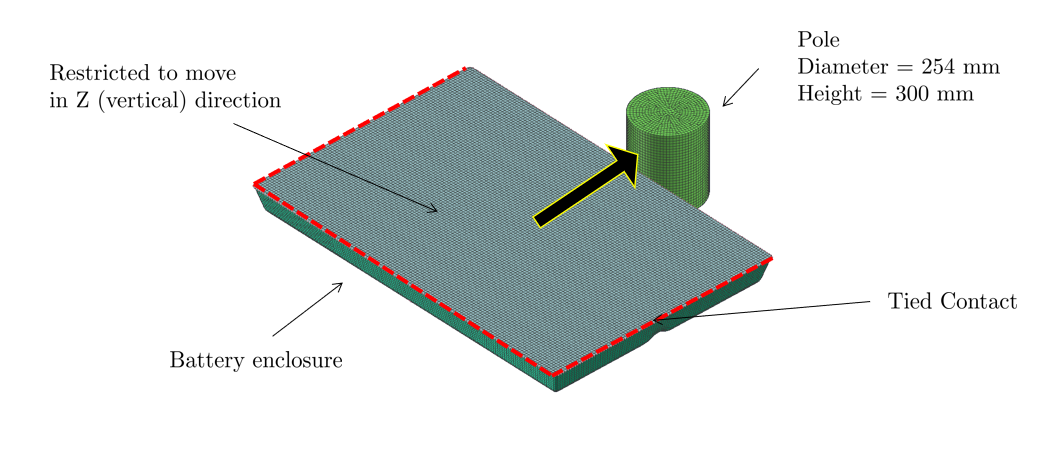}}
\caption{Schematic for thermoforming and crash simulation \cite{kulkarni2023investigation}}
\label{fig:schematic_thermo_crash}
\end{figure}

Later, the geometry of the die and the outline of the ply were imported into the software. Within the software, the punch geometry was created by replicating the die geometry and offsetting it by a distance equal to the total thickness of the composite sheet along the Z-direction. For example, in a case with $n_{ls}$ layers each of thickness $t_{l}$, the offset distance was $n_{ls} \times t_{l}$.

The composite layer was then created from the outline of the ply and subsequently meshed with PAMFORM's MAT140, a 4-node two-dimensional shell element, with a mesh size of about 3 mm and specified fiber orientations $\mathbf{\Phi}_{fib}$. The material properties of the ply were applied using the software's inbuilt material database \cite{pamformuser2019}. The same type of shell element and mesh size were used in all other thermoforming simulations.

The contact between all composite plies and between the die/punch and the composite layer was defined using a 'friction penalty.' Additionally, the coefficient of friction for this penalty was calculated by interpolating viscosity curves of the resin material, which is a function of shear rate and temperature. In PAMFORM, this functionality is implemented using look-up tables based on experimentally obtained data \cite{pamformuser2019}. The thermal conductivity and layer separation stress values used were 0.45 W/m·K and 0.005 GPa, respectively.

Both the punch and die were considered rigid and maintained at a temperature $T_{pd}$. The die was locked at the origin of the global reference axis, and the punch was allowed to move along the negative Z-direction at a velocity of $v_p$. The composite layer was kept at an initial temperature $T_i$, and the air temperature was maintained at $T_{air}$. The composite sheet was pressed against the die with a downward motion of the punch until it completely conformed to the shape of the die, at which point the simulation was terminated. The simulation is considered complete when the solver terminates without errors.

In the thermoforming process, composite sheets undergo significant deformation, resulting in the distortion of fiber orientations at specific locations. These alterations in fiber orientations plays an important role in determining the overall performance of the formed part. Furthermore, to recognize their significance, it is crucial to accurately incorporate these changes into the subsequent simulation phase, thereby ensuring the completeness of the simulation chain. This critical transition is known as draping, which functions as a link between the deformed state of the composite sheet after thermoforming and the next simulation phase, i.e. side pole impact crash simulation.

\subsubsection{Crash simulation setup}

A schematic of the side pole impact crash simulation is shown in Figure \ref{fig:schematic_thermo_crash} (b). Initially, the geometry was imported and meshed using VPS's shell MAT131, a 4-node two-dimensional shell element with one integration point, employing a mesh size of 5 mm. Subsequently, the result file from the thermoforming simulation, denoted by the extension ".erfh5", was imported into the software. The enclosure geometry was draped with this result file to update the material model, incorporating composite properties of the formed part.

Within the VPS, a stationary pole following the dimensions employed in actual side impact crash testing was created. Additionally, ribs and a lid were created within the software and connected to the enclosure geometry using a tied contact. This makes sure that the entire enclosure assembly functions as a unified entity throughout the impact. In terms of material properties, the pole was treated as rigid, while the ribs and lid were characterized as elasto-plastic. The material properties for the rib, lid, and enclosure used in the simulations are listed in the table in the Appendix.

To emulate realistic side pole impact conditions, the enclosure assembly was propelled toward the stationary rigid pole with a velocity of 32 km/h. The contact interaction between the enclosure and the pole was defined using the "Symmetric Node-to-Segment with Edge Treatment" card definition in VPS. The associated parameters, including a coefficient of friction of 0.2, stiffness proportional damping of 0.1, and a contact thickness equal to half of the laminate thickness, were employed. The simulation was performed for 10 milliseconds, with output data saved at intervals of 0.01 milliseconds.

\subsection{Automation pipeline}

The design variables, which encapsulate both the composite properties and processing conditions, are detailed in Table \ref{tab:doe_matrix}. These variables were selected by considering both commonly used material properties and realistic processing conditions typical of actual thermoforming processes.

\begin{table}[ht]
  \caption{Range used for creating DOE matrix}
  \centering
  \begin{tabular}{|c|c|} 
    \hline
    \textbf{Design Variables} & \textbf{Range} \\ 
    \hline
    \hline
    Number of layers, $n_{ls}$  & 4 - 16, only even \\ 
    \hline
    Thickness of each layer, $ t_{l}$ & 0.1 - 0.6 mm \\ 
    \hline
    Fiber orientations,  $\mathbf{\Phi}_{fib}$ & (0, 45, -45, 90), (30, -30, 60, -60) \\ 
    \hline
    Punch velocity,  $v_{p}$  & 4 - 6.5 m/s \\ 
    \hline
    Layer initial temperature,  $T_i$  & 200 - 400 \textdegree{}C \\ 
    \hline
    Punch / Die temperature, $T_{pd}$  & 20 - 220 \textdegree{}C  \\ 
    \hline
    Air temperature,  $T_{air}$ & 10 - 30 \textdegree{}C  \\
    \hline
  \end{tabular}
  \label{tab:doe_matrix}
\end{table}

Following these ranges, approximately 400 data points were sampled using Latin Hypercube Sampling to assure uniform coverage of the design space due to its space-filling properties \cite{iman2014atin, gramacy2020surrogates}.

\begin{figure}[h]
    \centering
    \includegraphics[width = 1\linewidth]{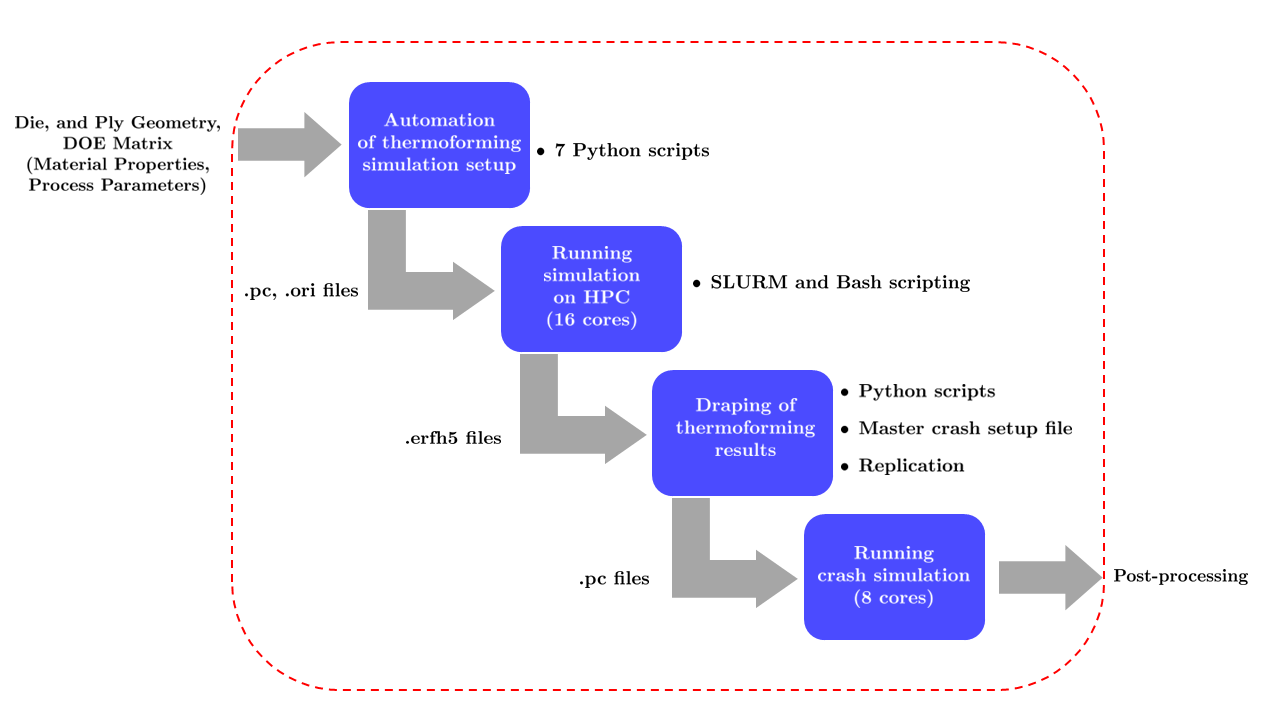}
    \caption{Schematic of a automation pipeline}
    \label{fig:schem_automat}
\end{figure}

Automation of the simulation chain, as shown in Figure \ref{fig:schem_automat}, consisted of four distinct stages. Initially, Python scripts were created to automate the thermoforming simulation setup, as detailed in Section \ref{sec:thermo}. These scripts generated simulation setup files for various layer configurations (e.g., configurations of four layers, six layers, etc.). The input parameters included die geometry, specific material, and processing parameters obtained from the DOE matrix. The scripts produced files with ".pc" and ".ori" extensions, containing simulation-specific details recognized by the solver.

Advancing to the second stage, the generated .pc files were uploaded to an HPC cluster. Subsequently, four simulations were run concurrently, each utilizing 16 cores to maximize throughput. Job scheduling on the HPC cluster was managed using SLURM scripting with job array functionality and supplemented with bash scripting. With this setup, a 10-layer thermoforming simulation, for example, required approximately 16 hours to complete

In the third stage, Python scripting was used to automate the generation of crash simulation setup files, leveraging the successful results of the thermoforming simulations. This process involved using the results from the thermoforming simulations to perform draping on the enclosure geometry of the master crash simulation setup file. Additionally, other parameters in the master crash simulation file that depended on draping were modified to create new crash simulation setup files.

Finally, the crash simulations were executed using the VPS solver, with each simulation running on eight cores at the same velocity and duration. The results from the crash simulations were later post-processed to extract the final values of the crash parameters. In total, 400 simulations were conducted, of which approximately 65\% were successful. This success rate is attributed to the inherent uncertainties in achieving the formability of a composite sheet through the thermoforming process, given a specific set of input parameters.



\subsection{Post-processing}



Following the simulation, Python scripts were developed to extract both the contact force plot between the enclosure and the pole, and the total energy absorbed by the enclosure assembly during the impact, saving these as CSV files. To calculate the intrusion of the stationary pole into the enclosure assembly, two edge sensor nodes were strategically positioned: one at the point of contact between the enclosure and the pole, and the other at the opposite end of the enclosure along the same line. Translation data from these sensor nodes during the impact were recorded and extracted as CSV files.

Using the extracted CSV files, crash parameters such as peak load ($F_p$), crush load efficiency ($\mathrm{CLE}$), specific energy absorption ($\mathrm{SEA}$), and deformation ($\mathrm{\Delta Y_{node}}$) were calculated. Each of these parameters is discussed in detail as follows:

\begin{itemize}
    \item \textbf{Peak Load ($F_p$):} This parameter represents the maximum load experienced by the enclosure during side pole impact \cite{ren2018progressive} \cite{singh2021}. It is obtained directly by selecting the maximum value of the contact force plot. It is a crucial measure to understand the highest force that the enclosure needs to withstand to prevent damage to the battery module inside.

    \item \textbf{Crush Load Efficiency ($\mathrm{CLE}$):} Calculated as the ratio of the average load during a side pole impact ($F_{avg}$) to the peak load ($F_p$) i.e. \begin{equation}
        \mathrm{CLE} = \frac{F_{avg}}{F_p}
    \end{equation} where,\begin{equation} F_{\mathrm{avg}}=\frac{\int_0^T F d t}{T}.
\end{equation} This metric provides insights into the stability of impact \cite{lukaszewicz2013automotive} \cite{Xu2016}\cite{Kim2011}. Its value ranges from 0 to 1 with 1 being theoretical maximum. A higher CLE value implies a stable impact.

    \item \textbf{Specific Energy Absorption ($\mathrm{SEA}$):} This is the most important parameter used in the automotive industry to evaluate the crashworthiness \cite{lukaszewicz2013automotive}\cite{Xu2016}\cite{Liu2014}. It provides a measure of how effectively the enclosure absorbs energy per unit its mass. Further, it given by dividing the energy absorbed ($\mathrm{EA}$) during an impact by the total mass of the enclosure assembly ($m_{\mathrm{total}}$). It is given by : \begin{equation}
        \mathrm{SEA} = \dfrac{\mathrm{EA}}{m_{\mathrm{total}}}.
    \end{equation}

    \item \textbf{Deformation ($\mathrm{\Delta Y_{node}}$):} This additional parameter was added to quantify penetration of pole inside the enclosure after impact. Understanding this is critical for assessing the structural integrity and safety of the battery pack. Ideally a composite enclosure should hinder the intrusion to protect the battery module inside. In this study, it was calculated as the displacement of a specific node on the enclosure after impact.
    
\end{itemize}

\section{Results and Discussion}\label{sec:others}

In the study, the initial step involved data pre-processing and cleaning, as a result, approximately 8\% of data points exhibiting high values of SEA were removed from the the dataset. The cleaned dataset included a total of 245 data points, and the input and output features were standardized using Z-scoring technique. Further, the dataset was divided into training (80\%) and testing (20\%) subsets. To evaluate the performance of ML model metrics such as $R^2$, which measures the goodness of fit, along with Mean Absolute Error (MAE) and Root Mean Squared Error (RMSE) were employed.

\subsection{Predictions on the holdout set}

The analysis began with fitting a linear regression (LR) model as a baseline. The LR model performed satisfactorily in predicting $F_p$ and $\mathrm{\Delta Y_{\text{node}}}$. However, it failed to accurately predict two other labels. To reduce the influence of irrelevant features, the LR model was regularized using $L_1$ (LASSO) and $L_2$ (Ridge) norms. The optimal regularization parameter $\lambda$ was determined through a grid search with five-fold cross-validation. It was found that LASSO with $\lambda = 0.01$ provided a marginal improvement, as shown in Table \ref{tab:results_lasso_GPR}, but the predictions for $\mathrm{SEA}$ and $\mathrm{CLE}$ remained subpar. This indicates that while linear regression is effective for certain types of data, it is not suitable for capturing more complex relationships or non-linear patterns.

\begin{table}[h]
    \centering
    \caption{Comparative results for performance of LASSO and GPR with different kernels}
    \footnotesize 
    \begin{tabular}{|c|c|c|c|c|c|c|c|c|c|c|c|c|}
    \hline
    Op / Est & \multicolumn{4}{c|}{LR (LASSO) ($\lambda = 0.01$)} & \multicolumn{4}{c|}{GPR (RBF)} & \multicolumn{4}{c|}{GPR (M32)} \\ 
    \hline
     & $R^2_{\text{train}}$ & $R^2_{\text{test}}$ & MAE & RMSE & $R^2_{\text{train}}$ & $R^2_{\text{test}}$ & MAE & RMSE &  $R^2_{\text{train}}$ & $R^2_{\text{test}}$ & MAE & RMSE  \\
    \hline
   $F_p$ & 0.952 & 0.954 & 13.971 & 19.238 & 0.981 & 0.959 & 13.71 & 18.146 & 0.988 & 0.962 & 13.267 & 17.502 \\ 
    \hline
   $\mathrm{CLE}$ & 0.668 & 0.263 & 0.019 & 0.024 & 0.867 & 0.481 & 0.016 & 0.02 & 0.908 & 0.474 & 0.016 & 0.02  \\
    \hline
    $\mathrm{SEA}$ & 0.441 & 0.338 & 2.486 & 3.322 & 0.758 & 0.512 & 2.076 & 2.854 & 0.847 & 0.553 & 2.021 & 2.731 \\
    \hline
     $\mathrm{\Delta Y_{node}}$ & 0.948 & 0.913 & 0.202 & 0.29 & 0.984 & 0.959 & 0.137 & 0.199 & 0.990 & 0.957 & 0.137 & 0.206 \\
    \hline
    \end{tabular}

    \label{tab:results_lasso_GPR}
\end{table}

Subsequently, we employed GPR using the GPy library to address the limitations of linear regression. The GPR models were constructed using various kernels, starting with the Radial Basis Function (RBF) kernel and progressing to other kernels such as Exponential, Matérn$\frac{5}{2}$, and Matérn$\frac{3}{2}$

The prediction results for the RBF and Matérn$\frac{3}{2}$ kernels are shown in Table \ref{tab:results_lasso_GPR}. As observed, the RBF kernel significantly improved the prediction of $\mathrm{SEA}$ and $\mathrm{CLE}$ compared to the LASSO regression. Additionally, the Matérn$\frac{3}{2}$ kernel marginally improved predictions.

\begin{figure}[h]
    \centering
    \includegraphics[width = 1\linewidth]{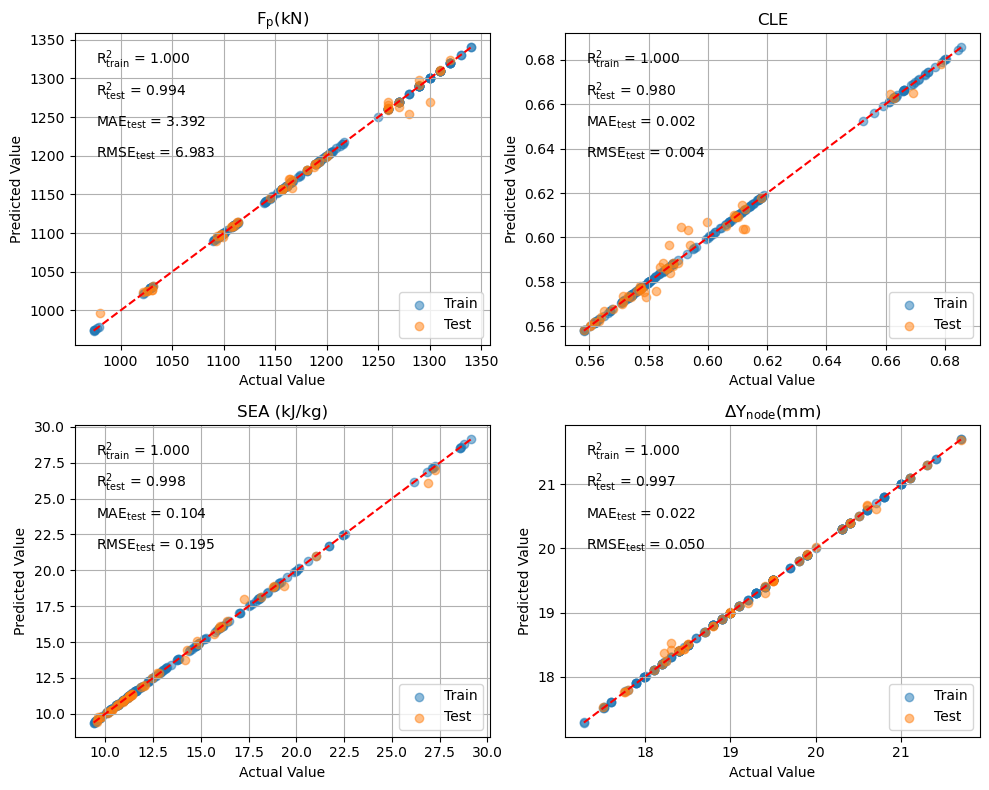}
    \caption{Predictions from GPR with Matérn$\frac{3}{2}$ + ARD kernel}
    \label{fig:predict_GPR}
\end{figure}

\begin{table}[h]
    
    \centering
    \caption{Comparative results for performance of GPR with different ARD kernels}
    \footnotesize 
    \begin{tabular}{|c|c|c|c|c|c|c|c|c|c|c|c|c|}
    \hline
    Op / Est & \multicolumn{4}{c|}{GPR (RBF + ARD)} & \multicolumn{4}{c|}{GPR (Exp + ARD)} & \multicolumn{4}{c|}{GPR (MAT32 + ARD)} \\ 
    \hline
    
      & $R^2_{\text{train}}$ & $R^2_{\text{test}}$ & MAE & RMSE & $R^2_{\text{train}}$ & $R^2_{\text{test}}$ & MAE & RMSE &  $R^2_{\text{train}}$ & $R^2_{\text{test}}$ & MAE & RMSE  \\
        
    \hline
   $F_p$ & 1 & 0.981 & 5.100 & 12.414 & 1 & 0.989& 3.518 & 9.189 & 1 & 0.994 & 3.392 & 6.983\\ 
    \hline
   $\mathrm{CLE}$ & 1 & 0.982 & 0.002 & 0.004 & 1 & 0.981 & 0.002 & 0.004 & 1 & 0.980 & 0.002 & 0.004  \\
    \hline
    $\mathrm{SEA}$ & 1 & 0.984 & 0.199 & 0.515 & 1 & 0.998 & 0.102 & 0.199  & 1 & 0.998 & 0.104 & 0.195 \\
    \hline
     $\mathrm{\Delta Y_{node}}$ & 1 & 0.994 & 0.040 & 0.075 & 1 & 0.996 & 0.023 & 0.063 & 1 & 0.997 & 0.022 & 0.05 \\
    \hline
    \end{tabular}
    
    \label{tab:results_GPR_ARD}
\end{table}

\begin{table}[h]
    \centering
     \caption{Comparative results for the performance of GPR, RF, GB, and XGBoost from the previous study \cite{shaikh2023finite} on the holdout set}
    \begin{tabular}{|c|c|c|c|c|c|c|c|c|}
    \hline
    Outputs & \multicolumn{2}{c|}{\shortstack{This study \\ (GPR)}} & \multicolumn{2}{c|}{\shortstack{RF}} & \multicolumn{2}{c|}{\shortstack{GB}} & \multicolumn{2}{c|}{\shortstack{XGBoost}} \\
    \hline
     & $R^2$ & MAE & $R^2$ & MAE & $R^2$ & MAE & $R^2$ & MAE \\
    \hline
    $F_p$ & 0.994 & 3.392 & - & - & - & - & - & - \\
    \hline
    $\mathrm{CLE}$ & \textbf{0.980} & \textbf{0.002} & 0.944 & 0.004 & 0.969 & 0.003 & 0.968 & 0.004 \\
    \hline
    $\mathrm{SEA}$ & 0.998 & 0.104 & - & - & - & - & - & - \\
    \hline
    $\mathrm{\Delta Y_{node}}$ & \textbf{0.997} & \textbf{0.022} & 0.967 & 0.109 & 0.970 & 0.086 & 0.969 & 0.094 \\
    \hline
    \end{tabular}
   
    \label{tab:GPR_RF_GB_XgBoost_Comparison}
\end{table}

Finally, to improve the predictions further, automatic relevance determination (ARD) \cite{liu2019gaussian} was incorporated in each tested kernels. ARD allows models to adaptively adjust the relevance of input features to improve predictions. This is discussed in detail in section \ref{sec:gpr}. 

As detailed in Table \ref{tab:results_GPR_ARD}, each kernel used previously was tested with ARD. We found that the Matern$\frac{3}{2}$ with ARD provided excellent results in achieving the highest $R^2$ scores across all output labels, as seen in figure \ref{fig:predict_GPR}. Further, it can be noted that Matern$\frac{3}{2}$ kernel with ARD was able to effectively capture the underlying patterns in the dataset and to produce more accurate predictions compared to other models

Performance of the GPR model was compared against the outcomes reported in previous study using tree-based ensemble methods \cite{shaikh2023finite}. The primary objective was to assess the improvements in predictive accuracy and error reduction achieved through the application of GPR model.

Table \ref{tab:GPR_RF_GB_XgBoost_Comparison} presents a summarized comparison of the performance metrics between the two studies. Notably, GPR provided a significant reduction in error for output variables such as $\mathrm{CLE}$ and $\mathrm{\Delta Y_{node}}$. The higher $R^2$ values and substantially lower MAE with GPR suggest superior ability in modeling complex patterns within the dataset, resulting in an improved predictions. 

\subsection{Repeated K-Fold Cross Validation}

In this work, we employed repeated K-fold cross-validation to assess the consistency and degree of generalization of the GPR model. This technique involves dividing the dataset into $k$ equally sized subsets and iteratively using $k-1$ of these for training the model, while the remaining subset is reserved for testing.

Repeating this process multiple times with shuffled data splits assures the reliability of performance metrics by reducing the impact of potential randomness in the initial partitioning. This approach provides a more comprehensive view of model performance, thus making sure the predictions are robust and reflective of the model's ability to generalize to diverse data instances.

\begin{figure}[h]
    \centering
    \includegraphics[width = 0.8 \linewidth]{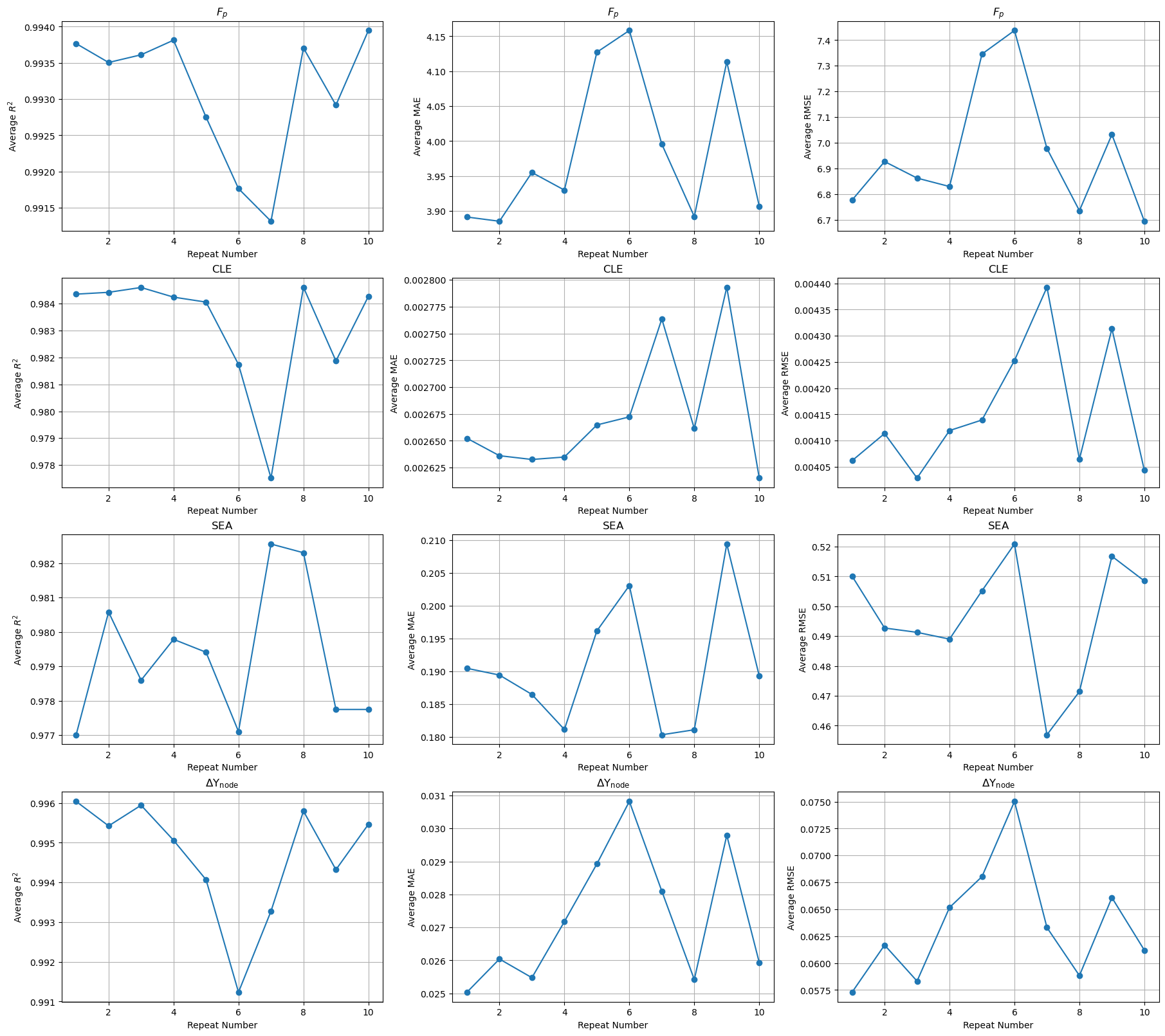}
    \caption{Average values of $R^2$, MAE, RMSE across different repeats for fivefold cross validation repeated 10 times }
    \label{fig:rf_average}
\end{figure}

Figure \ref{fig:rf_average} depicts the average values of the performance metrics—$R^2$, MAE, and RMSE—over 10 repeats. For each output—$F_p$, $\mathrm{CLE}$, $\mathrm{SEA}$, and $\Delta Y_{\text{node}}$—the variation in performance metrics across the 10 repeats is evident. This variability is a normal aspect of repeated K-fold cross-validation, reflecting how different splits of the data can influence the model's performance.

Despite the fluctuations, the graphs generally indicate a strong predictive capability, as shown by the high $R^2$ values, and low MAE and RMSE values, which is indicative of a well-performing model. These results underscore the robustness of the GPR model, as it consistently yields reliable predictions across different subsets of data. 

\subsection{Prediction on new instances}

Following the evaluation of the model with repeated K-fold cross-validation in the preceding section, we further assessed the performance of the GPR model using a new dataset that consisted of 10 data points generated through new simulations. In each simulation, $n_{ls} = 4$, while other properties such as $v_{p}$, $T_i$, $T_{pd}$, and $T_{air}$ were randomly selected from Table \ref{tab:doe_matrix}. The values of $t_l$ and $\mathbf{\Phi}_{fib}$ were set outside the previously used range \cite{wang2016effects}.

The thermoforming simulations were performed on a HPC unit with 16 cores, while the crash simulations were conducted on an 8-core 11\textsuperscript{th} Gen Intel (R) i7@3GHz PC. The total time taken for one complete simulation was approximately 29,000 seconds (i.e., 8 hours).

\begin{table}[h]
    \centering
    \caption{$n_{ls}$ = 4; $v_{p}$, $T_i$, $T_{pd}$, $T_{air}$ randomly selected (from table \ref{tab:doe_matrix}); $t_{l}$ and $\mathbf{\Phi}_{fib}$, different from testing and training }
    \begin{tabular}{|c|c|c|c|c|c|c|}
     \hline
      Outputs & \multicolumn{3}{c|}{\shortstack{0.7 mm, 6.5 m/s,\\ 318 °C, 131 °C , 24 °C,\\ (45, -45, 45, -45)}} & \multicolumn{3}{c|}{\shortstack{0.9 mm, 5.7 m/s,\\ 304 °C, 91 °C , 22 °C, \\ (45, -45, 45, -45)}}\\
      \hline
       & Sim. & GPR & \% error & Sim. & GPR & \% error \\
      \hline
      $F_p$ & 1050.0 & 1040.7 & \textbf{0.88} & 1030.0 & 1092.4 & \textbf{6.06} \\
      $\mathrm{CLE}$ & 0.527 & 0.577 & \textbf{9.43} & 0.542 & 0.581 & \textbf{7.41} \\
      $\mathrm{SEA}$ & 13.61 & 12.18 & \textbf{10.51} & 13.78 & 14.25 & \textbf{3.44} \\
      $\mathrm{\Delta Y_{node}}$ & 17.14 & 17.52 & \textbf{2.24} & 16.85 & 17.84 & \textbf{5.90} \\
      \hline
    \end{tabular}
    \label{tab:newdataseta}
\end{table}

For the sake of brevity, the results of only four simulations are detailed in Tables \ref{tab:newdataseta} and \ref{tab:newdatasetb}. The results for the remaining simulations are listed in the appendix. It is important to note that significantly different values of $t_l$ and $\mathbf{\Phi}_{fib}$ were used in these simulations compared to those used in the data generation step. For comparison, the predictions from the GPR model are listed alongside the simulation results.

Additionally, it should be noted that the GPR model provided approximate predictions instantaneously, compared to the actual simulations. The percentage errors for the four simulations were well within a reasonable range, with the lowest being 0.88\% for $F_p$ and the highest being 14.64\% for $\mathrm{SEA}$. This demonstrates the predictive accuracy of the GPR model, with its predictions consistently close to the simulation results. This further underscores its effectiveness in capturing the complex relationships within the dataset.

\begin{table}[h]
    \centering
     \caption{$n_{ls}$ = 4; $v_{p}$, $T_i$, $T_{pd}$, $T_{air}$  and $\mathbf{\Phi}_{fib}$ randomly selected (from table \ref{tab:doe_matrix}); $t_{l}$ different from training and testing }
    \begin{tabular}{|c|c|c|c|c|c|c|}
     \hline
      Outputs & \multicolumn{3}{c|}{\shortstack{0.7 mm, 5.8 m/s,\\ 239 °C, 68 °C , 16 °C,\\ (0, 45, -45, 90)}} & \multicolumn{3}{c|}{\shortstack{0.8 mm, 4.9 m/s,\\ 206 °C, 219 °C , 25 °C,\\ (0, 45, -45, 90) }}\\
      \hline
       & Sim. & GPR & \% error & Sim. & GPR & \% error \\
      \hline
      $F_p$ & 974.0 & 1017.6 & \textbf{4.43} & 971.0 & 1051.4 & \textbf{8.28} \\
      $\mathrm{CLE}$ & 0.562 & 0.579 & \textbf{3.11} & 0.569 & 0.578 & \textbf{1.74} \\
      $\mathrm{SEA}$ & 14.79 & 12.62 & \textbf{14.64} & 14.46 & 13.46 & \textbf{6.90} \\
      $\mathrm{\Delta Y_{node}}$ & 16.33 & 17.29 & \textbf{5.86} & 16.34 & 17.50 & \textbf{7.11} \\
      \hline
    \end{tabular}
    \label{tab:newdatasetb}
\end{table}

Subsequently, further analysis of results from the entire dataset revealed that the mean absolute percentage error for $F_p$, $\mathrm{CLE}$, $\mathrm{SEA}$, and $\mathrm{\Delta Y_{node}}$ was 4.09\%, 5.76\%, 8.08\%, and 5.48\%, respectively. Additionally, the percentage error for all response variables across all simulations was less than or equal to 14.64\%. The variation in error rates across different outputs and configurations underscores the model's sensitivity to input parameters.

\subsection{Uncertainty estimate from the GPR posterior}

The strength of GPR lies in its ability to provide both predictions and associated uncertainty, indicative of the model's confidence in its predictions. The mean of the GPR posterior represents the predictions, while the variance, which is used to calculate the confidence interval, provides an estimate of uncertainty.

\begin{figure}[h]
    \centering
    \includegraphics[width = 1\textwidth]{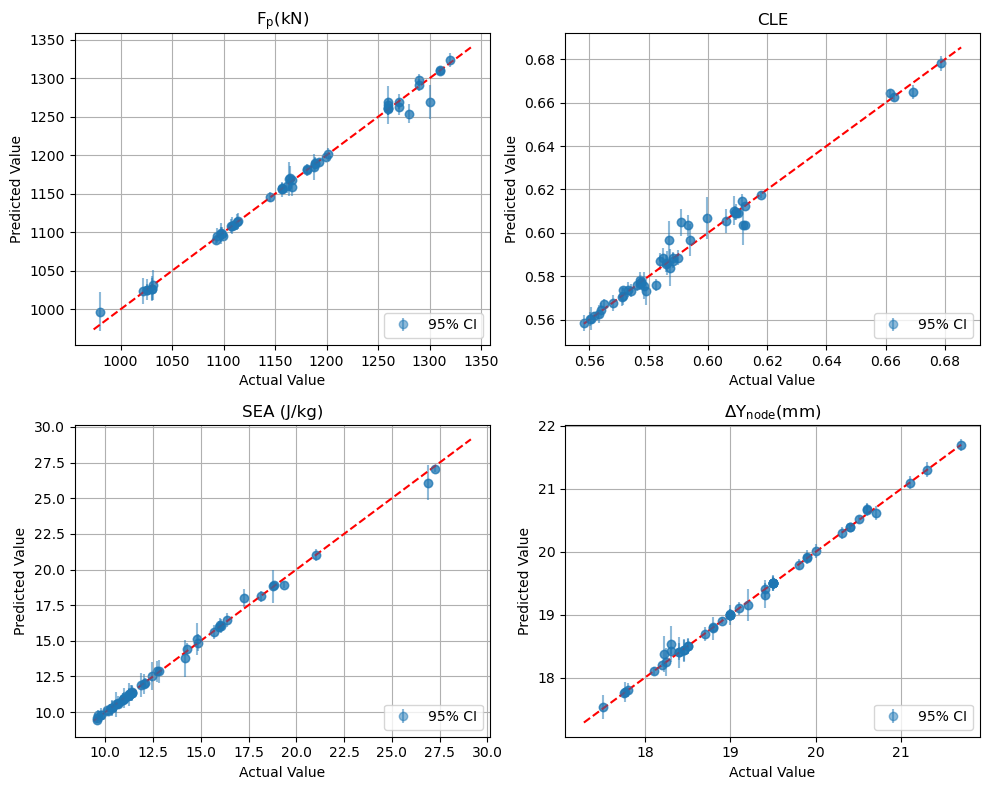}
    \caption{Uncertainty estimate (95\% confidence interval) from GPR (Matérn$\frac{3}{2}$ + ARD) posterior on holdout set represented as an error bar}
    \label{fig:GPR_UQ}
\end{figure}

The uncertainty estimates from the GPR model using the Matérn$\frac{3}{2}$ kernel with ARD for $F_p$, $\mathrm{CLE}$, $\mathrm{SEA}$, and $\mathrm{\Delta Y_{node}}$ are shown in Figure \ref{fig:GPR_UQ}. The blue error bars represent a 95\% confidence interval, indicating the range within which the actual value is expected to lie with 95\% probability, assuming a normal distribution of the predictions.

From the plots, it is evident that the confidence intervals for almost all output labels are narrow, implying the model's accuracy and high confidence in its predictions. However, it is important to note that this estimate primarily represents the epistemic component of total uncertainty \cite{xu2021pitfalls} \cite{hoffer2022gaussian}. Because inputs to the model were considered without uncertainty and the noise variance estimate obtained from GPR is close to zero, the absence of an aleatoric component is suggested.

Propagating the input uncertainties within the GPR framework increases modeling complexity as this approach challenges the assumption of a Gaussian distributed posterior \cite{bijl2016gaussian} \cite{mchutchon2011gaussian}. Specifically, including input uncertainty introduces additional variability that may affect the shape of the resulting posterior distribution, potentially deviating from the Gaussian form \cite{girard2004approximate}. To address this challenge, several techniques such as Taylor series expansion, moment matching integration are employed. A comprehensive list of techniques used to tackle this problem are enumerated here \cite{Johnson2023UncertainGP}. 

Given the inherent complexity of performing uncertainty propagation within the GPR framework, this study employed Monte Carlo uncertainty propagation as described in the next section.

\subsection{Monte Carlo uncertainty propagation}

We conducted the Monte Carlo uncertainty propagation study to evaluate the impact of input variable uncertainties on outputs. These uncertainties were identified with device measurement errors in mind, reflecting the conditions encountered in the realistic thermoforming process. The simulation scenarios were established by dividing the range outlined in Table \ref{tab:doe_matrix} into three distinct cases: two at the boundaries of the range and one at the midpoint. For each scenario, two fiber orientation configurations were analyzed: Configuration A, defined by $\mathbf{\Phi}_{fib} = [0, 45, -45, 90]$, and Configuration B, defined by $\mathbf{\Phi}_{fib} = [30, -30, 60, -60]$.

The input variables $t_{l}$, $v_{p}$, $T_i$, $T_{pd}$, and $T_{air}$ were assumed to follow a normal distribution with probability distribution parameters listed in Table \ref{tab:prob_mcuq}. Additionally, the input variable $\mathbf{\Phi}_{fib}$, though not listed in the table, was also assumed to be normally distributed with $\mu = A/B$ (i.e., $[0, 45, -45, 90]$ or $[30, -30, 60, -60]$) and $\sigma = 2^{\circ}$ \cite{rahul2024statistical}. The uncertainty in the variable $n_{ls}$ was not considered because of its deterministic nature in a realistic manufacturing setup.

\begin{table}[h]
    \centering
     \caption{Probability distribution of input variables used for Monte Carlo uncertainty quantification study}
      \fontsize{10pt}{11pt}\selectfont
    \begin{tabular}{|>{\centering\arraybackslash}m{1cm}|>{\centering\arraybackslash}m{2cm}|>{\centering\arraybackslash}m{2cm}|>{\centering\arraybackslash}m{2cm}|>{\centering\arraybackslash}m{2cm}|}
     \hline
      Input Var. & \shortstack{Assumed \\ Prob.\\ Dist } & \multicolumn{3}{c|}{ Monte Carlo Simulation} \\
      \hline
      &  & Case \#1 & Case \#2 & Case \#3 \\
      \hline
       $n_{ls}$ & Const. & 4 & 10 & 16 \\
      
      $t_{l}$ & Normal & \shortstack{ $\mu = 0.1$,\\ $\sigma$ = 1\% $\mu$  } & \shortstack{ $\mu$ = 0.35,\\ $\sigma$ = 1\% $\mu$ } & \shortstack{ $\mu$ = 0.6,\\ $\sigma$ = 1\% $\mu$ } \\
     
      $ v_{p} $ & Normal  & \shortstack{ $\mu$ = 4,\\ $\sigma$ = 1\% $\mu$ } & \shortstack{ $\mu$ = 5.25,\\ $\sigma$ = 1\% $\mu$ } & \shortstack{ $\mu$ = 6.5,\\ $\sigma$ = 1\% $\mu$ }\\
     
      $ T_i $ & Normal  & \shortstack{ $\mu$ = 200,\\ $\sigma$ = 1\% $\mu$ } & \shortstack{$\mu$ = 300,\\ $\sigma$ = 1\% $\mu$} & \shortstack{$\mu$ = 400,\\ $\sigma$ = 1\% $\mu$} \\
       
      $T_{pd}$ & Normal & \shortstack{ $\mu$ = 20, \\ $\sigma$ = 0.75\% $\mu$ } & \shortstack{ $\mu$ = 120,\\ $\sigma$ = 0.75\% $\mu$ } & \shortstack{ $\mu$ = 220, \\ $\sigma$ = 0.75\% $\mu$ } \\
      
       $T_{air}$ & Normal & \shortstack{ $\mu$ = 10,\\ $\sigma$ = 1.5\% $\mu$ } & \shortstack{ $\mu$ = 20,\\ $\sigma$ = 1.5\% $\mu$ } & \shortstack{ $\mu$ = 30,\\ $\sigma$ = 1.5\% $\mu$ } \\
      \hline
    \end{tabular}
   
    \label{tab:prob_mcuq}
\end{table}

\begin{figure}[h]
        \centering
         \subfloat[Case \# 1]{\includegraphics[width=0.4\linewidth]{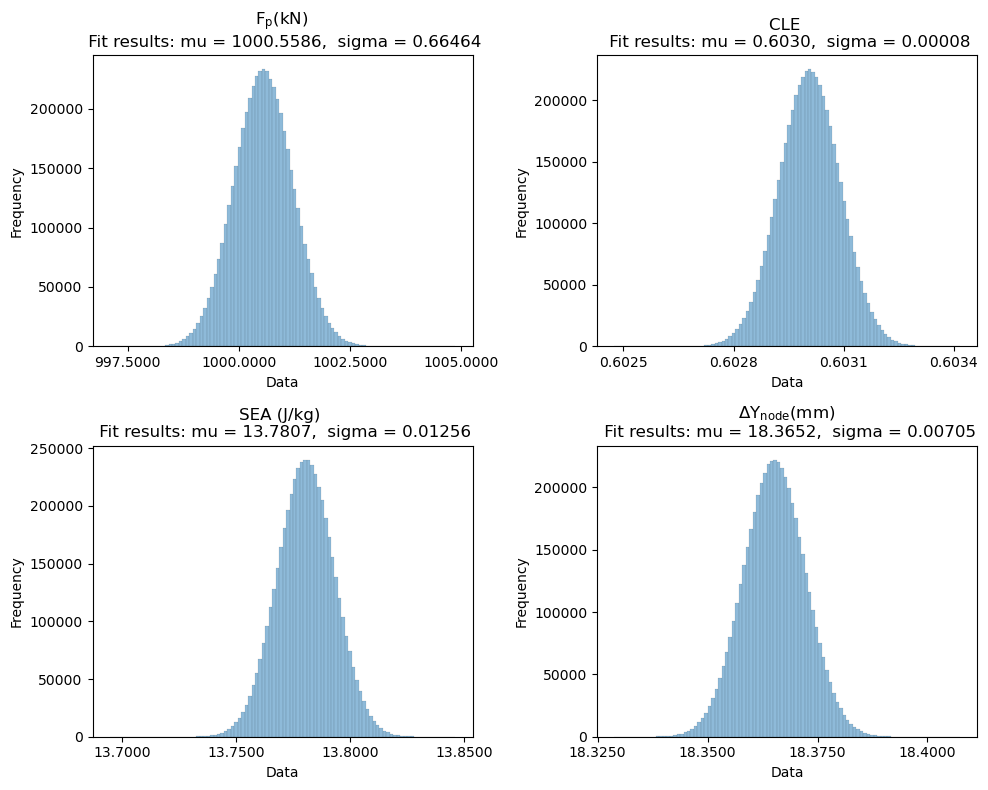}}
        \subfloat[Case \# 2]{\includegraphics[width=0.4\linewidth]{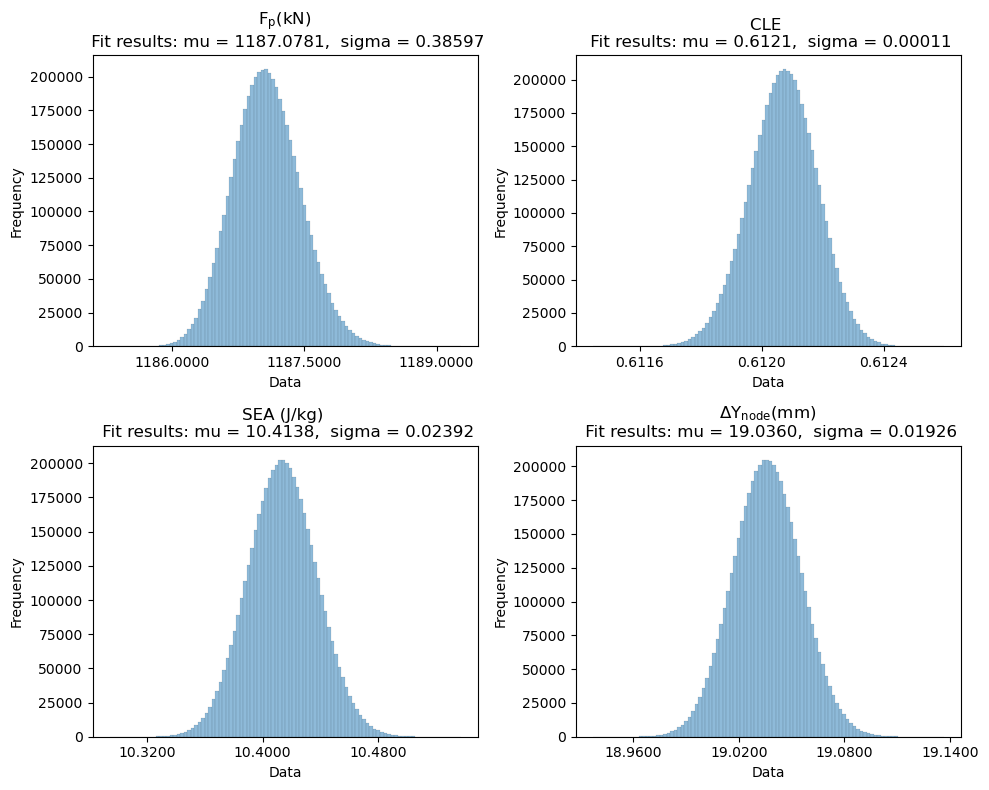}}\\
        \subfloat[Case \# 3]{\includegraphics[width=0.38\linewidth]{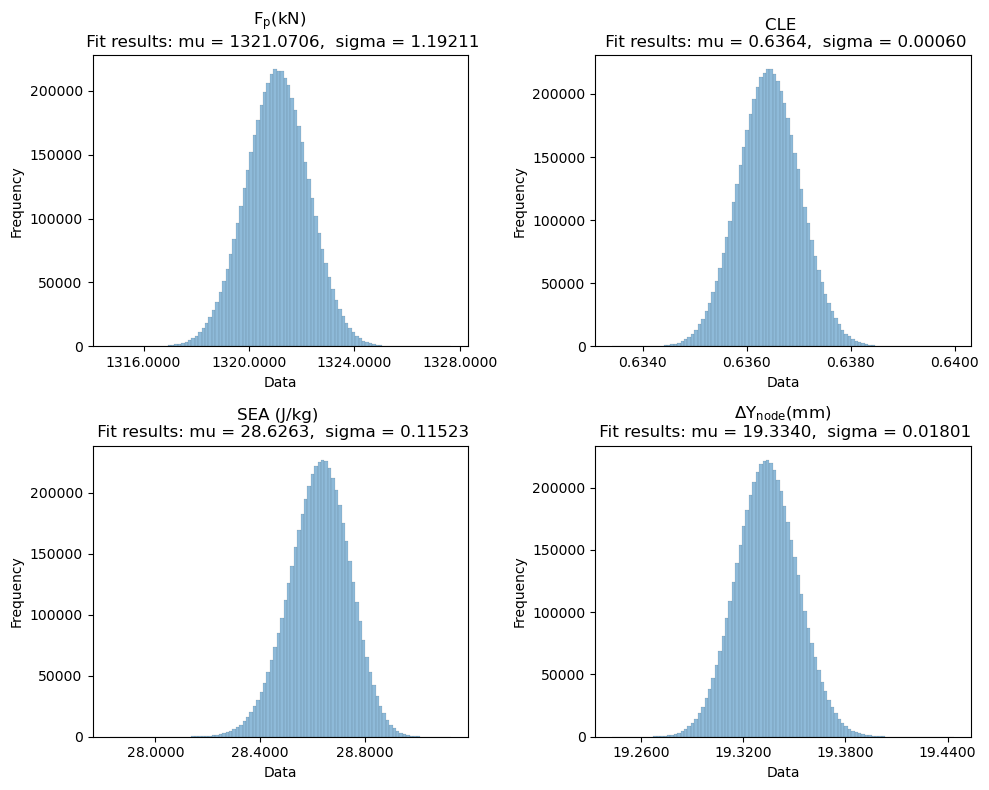}}
        \caption{Results from three cases with configuration A}
        \label{fig:UQ_results1}
    \end{figure}

\begin{figure}[h]
        \centering
         \subfloat[Case \# 1]{\includegraphics[width=0.4\linewidth]{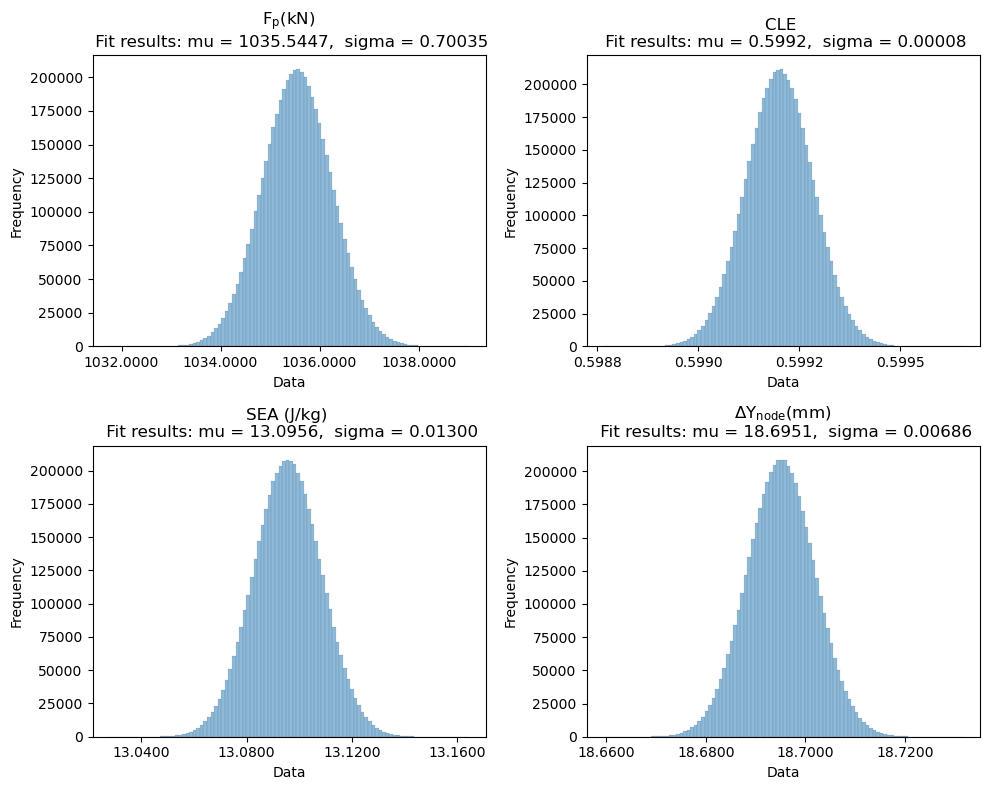}}
        \subfloat[Case \# 2]{\includegraphics[width=0.4\linewidth]{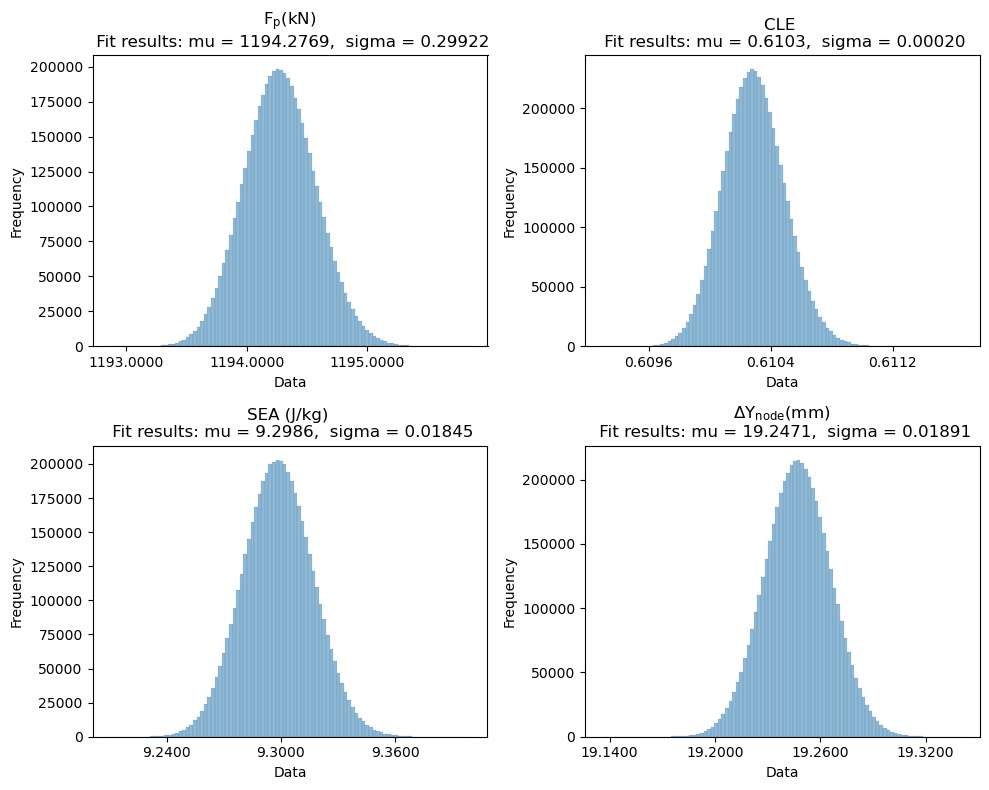}}\\
        \subfloat[Case \# 3]{\includegraphics[width=0.38\linewidth]{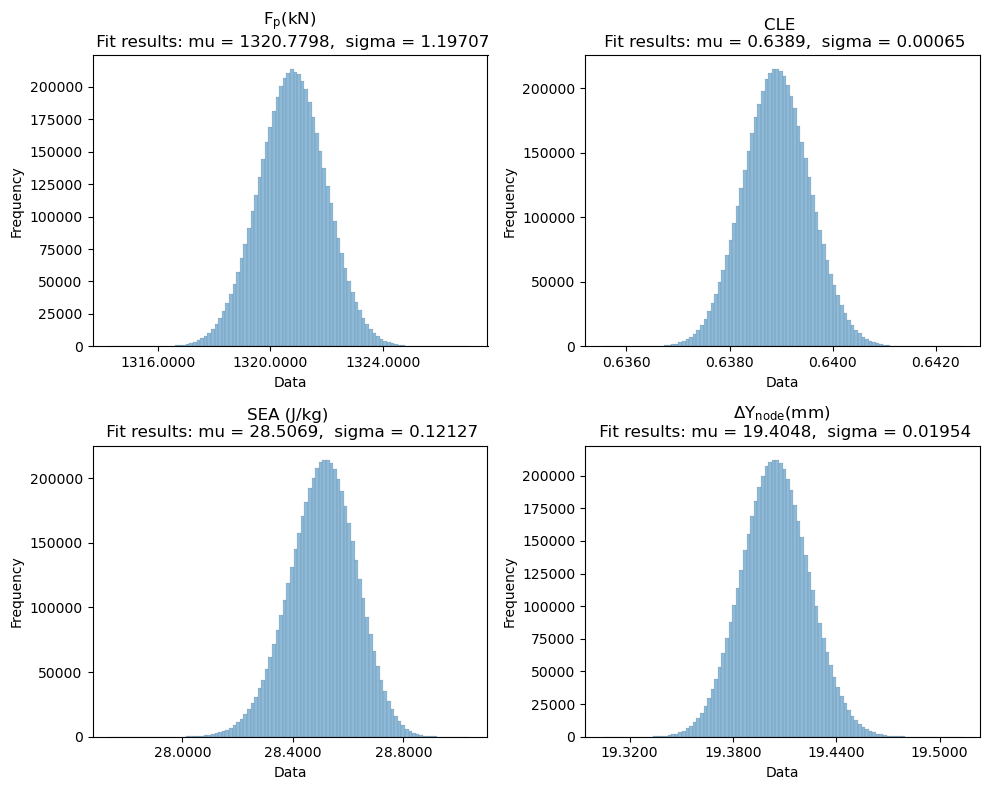}}
        \caption{Results from three cases with B configuration}
        \label{fig:UQ_results2}
    \end{figure}

We conducted a total of six MC simulations by sampling $5 \times 10^6$ samples and running the pre-trained GPR model on each sample. The results were aggregated, and statistical parameters such as the mean and standard deviation were extracted by fitting the observed distribution, as shown in Figures \ref{fig:UQ_results1} to \ref{fig:UQ_results2}.

\begin{table}[h!]
    \centering
    \caption{Mean and standard deviation (as \% of mean) for cases with configuration A}
    \begin{tabular}{lcccccc}
        \toprule
        \textbf{Output Label} & \multicolumn{2}{c}{\textbf{Case \#1}} & \multicolumn{2}{c}{\textbf{Case \#2}} & \multicolumn{2}{c}{\textbf{Case \#3}} \\
        \cmidrule(r){2-3} \cmidrule(r){4-5} \cmidrule(r){6-7}
        & \(\mu\) & \(\sigma\) as \% of \(\mu\) & \(\mu\) & \(\sigma\) as \% of \(\mu\) & \(\mu\) & \(\sigma\) as \% of \(\mu\) \\
        \midrule
        \( F_p \) (kN) & 1000.5586 & 0.066427 & 1187.0781 & 0.032514 & 1321.0706 & 0.090238 \\
        CLE & 0.6030 & 0.013267 & 0.6121 & 0.017971 & 0.6364 & 0.094280 \\
        SEA (J/kg) & 13.7807 & 0.091142 & 10.4138 & 0.229695 & 28.6263 & 0.402532 \\
        \( \mathrm{\Delta Y_{\text{node}}} \) (mm) & 18.3652 & 0.038388 & 19.0360 & 0.101177 & 19.3340 & 0.093152 \\
        \bottomrule
    \end{tabular}
    \label{tab:UQA}
\end{table}

Further, results are summarized in Tables \ref{tab:UQA} and \ref{tab:UQB}. The mean values $F_p$ range from 1000.5586 to 1321.0706, with standard deviations as a percentage of the mean varying from 0.025056\% to 0.090585\%. The CLE variable shows low variability with mean values between 0.5992 and 0.6389 and standard deviations from 0.013267\% to 0.101727\%. The SEA exhibits significant variability, particularly in Case \#3, with a standard deviation of 0.425327\%. Finally, $\Delta Y_{\text{node}}$ demonstrates moderate variability across cases, with mean values from 18.3652 to 19.4048 and standard deviations from 0.036704\% to 0.101177\%. 

\begin{table}[H]
    \centering
    \caption{Mean and standard deviation (as \% of mean) for cases with configuration B}
    \begin{tabular}{lcccccc}
        \toprule
        \textbf{Output Label} & \multicolumn{2}{c}{\textbf{Case \#1}} & \multicolumn{2}{c}{\textbf{Case \#2}} & \multicolumn{2}{c}{\textbf{Case \#3}} \\
        \cmidrule(r){2-3} \cmidrule(r){4-5} \cmidrule(r){6-7}
        & \(\mu\) & \(\sigma\) as \% of \(\mu\) & \(\mu\) & \(\sigma\) as \% of \(\mu\) & \(\mu\) & \(\sigma\) as \% of \(\mu\) \\
        \midrule
        \( F_p \) (kN) & 1035.5447 & 0.067625 & 1194.2769 & 0.025056 & 1320.7798 & 0.090585 \\
        CLE & 0.5992 & 0.013347 & 0.6103 & 0.032772 & 0.6389 & 0.101727 \\
        SEA (J/kg) & 13.0956 & 0.099269 & 9.2986 & 0.198376 & 28.5069 & 0.425327 \\
        \( \mathrm{\Delta Y_{\text{node}}} \) (mm) & 18.6951 & 0.036704 & 19.2471 & 0.098274 & 19.4048 & 0.100717 \\
        \bottomrule
    \end{tabular}
     \label{tab:UQB}
\end{table}

Across all simulations, the resulting histogram of outputs was observed to follow normal distribution suggesting that the input uncertainties propagated through the model resulting in normally distributed outcomes with predictable variance. Additionally, it was observed  that the variability in the results was relatively low, with standard deviations being a small fraction of the means.

\section{Conclusion}

This study began with the data generation step, which involved automating a complex finite element simulation chain for thermoforming and crash simulations. To streamline the process, we made several improvements to the simulation setup. These time- and resource-intensive simulations were performed using HPC clusters., and the results then were post-processed and assembled into a dataset. A GPR model was fitted, and the results were presented in the preceding section. Our analysis showed that the GPR model predicted output labels with high accuracy, achieving an $R^2$ > 0.980, MAE < 3.392, and RMSE < 6.983 for all output labels.

Additionally, we validated the GPR model's consistency and generalization using repeated K-fold cross-validation. The average evaluation metrics for different repeats were found to be close to the obtained predictions.

We also tested the model's predictive performance on the new dataset . The model provided predictions with an absolute percentage error of less than 14.64\% for all output labels. Furthermore, the mean absolute percentage error for all output labels on the entire dataset was less than 8.08\%. The model generated these approximate predictions in a few milliseconds, compared to the actual simulations which took about 29,000 seconds (i.e., 8 hours) each.

Finally, we conducted a Monte Carlo uncertainty propagation by incorporating uncertainty in the inputs and propagating it through the surrogate model. This process within the GPR framework adds complexity. We observed that normally distributed uncertainty in the input variables propagated through the GPR model, resulting in normally distributed predictions. Additionally, we found that the uncertainty in input variables had a small overall impact on the output labels.

In conclusion, the GPR model was able to capture the complex relationships within the dataset, providing output labels with higher predictive accuracy compared to the previous study \cite{shaikh2023finite}.

While this research has shown promise, several key areas have been identified for future exploration. This study focused on the performance of the enclosure in a side pole impact test; however, other modes of failure during a collision also should be considered for a comprehensive safety assessment. Additionally, the fiber volume fraction was kept constant; however, varying this parameter may affect the energy absorption capabilities of the composite structure and should be included in future analyses.

\section*{Acknowledgments}
The authors would like to acknowledge the help from Arnaud Dereims (ESI North America) and Ramesh Dwarampudi (ESI North America) regarding ESI software. Further, would like to extend their thanks to all colleagues at PNNL, especially Dr. Mohammad Faud Nur Taufique; and PNNL Research Computing for their technical support. Lastly, Mechanical Engineering and Engineering Science Department at the University of North Carolina Charlotte for their support.

\section*{Funding}
This work was supported by the High-Performance Computing for Energy Innovation program, managed by U.S. Department of Energy’s Office of Energy Efficiency and Renewable Energy under the proposal FP-D-20.1-23733. 

\bibliographystyle{unsrt}  
\bibliography{references}  

\appendix

\section*{Appendix}

\section{Material properties used in thermoforming and crash simulations}

\begin{table}[h!]
    \centering
    \caption{Material properties for the die and punch}
    \begin{tabular}{|l|l|}
     \hline
      Property & Value \\
      \hline
      Mechanical properties & Rigid material \\
      Convection coefficient & 10 W/m\(^2\) K \\
      Conductivity & 0.45 W/m.K \\
      \hline
    \end{tabular}
    \label{tab:material_properties_pd}
\end{table}

\begin{table}[h!]
    \centering
    \caption{Material properties of composite sheets}
    \begin{tabular}{|l|l|}
     \hline
      Property & Value \\
      \hline
      Density & \(1 \times 10^{-6}\) kg/mm\(^3\) \\
      Initial angle between fibers & 90 degrees \\
      Thickness & 0.25 mm \\
      Fiber content & 0.5 \\
      Tension compression stiffness (Fiber 1) & 20 GPa \\
      Tension compression stiffness (Fiber 2) & 20 GPa \\
      Bending stiffness (Fiber 1) & 0.03 GPa \\
      Bending stiffness (Fiber 2) & 0.03 GPa \\
      Conductivity & \(2.3 \times 10^{-6}\) kW/mm °C \\
      Specific heat & 1150 J/kg °C \\
      In plane shear & \(2.5 \times 10^{-5}\) GPa \\
      Sheet orientation & (90, 45-, 45, 0) \\
      Layer separation stress & 0.005 GPa \\
      Convection coefficient & 10 W/m\(^2\)K \\
      \hline
    \end{tabular}
    \label{tab:composite_material_specifications}
\end{table}

\begin{table}[h!]
    \centering
    \caption{Mechanical properties for the lid and rib}
    \begin{tabular}{|l|l|}
     \hline
      Property & Value \\
      \hline
      Density & \(1.8 \times 10^{-6}\) kg/mm\(^3\) \\
      Young's modulus & 125 GPa \\
      Yield stress & 3.5 GPa \\
      Poisson's ratio & 0.33 \\
      Max plastic strain for element removal & 0.014 \\
      Plastic tangent modulus & 8 GPa \\
      \hline
    \end{tabular}
    \label{tab:mechanical_properties_material_lid_rib}
\end{table}

\begin{table}[h!]
    \centering
    \caption{Material properties of enclosure used for crash simulations}
    \begin{tabular}{|l|l|}
     \hline
      Property & Value \\
      \hline
      Density & \(1.8 \times 10^{-6}\) kg/mm\(^3\) \\
      Young's modulus parallel to fiber & 125 GPa \\
      Young's modulus perpendicular to fiber & 8 GPa \\
      Critical shear damage limit & 0.114 GPa \\
      Initial shear damage limit & 0.02 GPa \\
      Initial strain of tensile fiber & 0.012 \\
      Ultimate strain of tensile fiber & 0.014 \\
      Tensile fiber ultimate damage & 0.99 \\
      Initial strain compressive fiber & 0.008 \\
      Ultimate strain compressive fiber & 0.009 \\
      Compressive fiber ultimate damage & 0.99 \\
      Initial yield stress & 0.02 GPa \\
      Hardening law exponent & 0.64 \\
      Hardening law multiplier & 1.3 \\
      Shear modulus 1,2 plane & 7 GPa \\
      Shear modulus 2,3 plane & 4 GPa \\
      Shear modulus 1,3 plane & 4 GPa \\
      Poisson's ratio & 0.33 \\
      Critical transverse damage limit & 1 \\
      Initial transverse damage limit & 0.02 \\
      \hline
    \end{tabular}
    \label{tab:detailed_material_properties_crash}
\end{table}

\newpage

\section{Predictions on remaining new dataset}

\begin{table}[h]
    \centering
    \begin{tabular}{|c|c|c|c|c|c|c|}
     \hline
      Outputs & \multicolumn{3}{c|}{\shortstack{0.5 mm, 5.8 m/s,\\ 239 °C, 68 °C , 16 °C,\\ (0, 45, -45, 60)}} & \multicolumn{3}{c|}{\shortstack{0.1 mm, 4.9 m/s,\\ 206 °C, 219 °C , 25 °C, \\ (0, 45, -45, 60)}}\\
      \hline
       & Sim. & GPR & \% err. & Sim. & GPR & \% err.\\
      \hline
      $F_p$ & 1010 & 976.2 & \textbf{3.35} & 1010 & 998.6 & \textbf{1.12}  \\
      $\mathrm{CLE}$ & 0.545 & 0.583 & \textbf{6.97} & 0.566 & 0.607 & \textbf{7.08} \\
      $\mathrm{SEA}$ & 14.17 & 12.91 & \textbf{8.88} & 15.46 & 13.73 & \textbf{11.16} \\
      $\mathrm{\Delta Y_{node}}$ & 16.72 & 17.31 & \textbf{3.50} & 16.81 & 18.39 & \textbf{9.40} \\
      \hline
    \end{tabular}
    \caption{$n_{ls}$ = 4; $t_{l}$, $v_{p}$, $T_i$, $T_{pd}$, $T_{air}$ randomly selected (from table \ref{tab:doe_matrix}); $\mathbf{\Phi}_{fib}$ different from training and testing}
    \label{tab:newdataset1}
\end{table} 

\begin{table}[h]
    \centering
    \begin{tabular}{|c|c|c|c|c|c|c|}
     \hline
      Outputs & \multicolumn{3}{c|}{\shortstack{0.7 mm, 5.8 m/s,\\ 239 °C, 68 °C , 16 °C,\\ (0, 45, -45, 90)}} & \multicolumn{3}{c|}{\shortstack{0.8 mm, 4.9 m/s,\\ 206 °C, 219 °C , 25 °C,\\ (0, 45, -45, 90) }}\\
      \hline
       & Sim. & GPR & \% error & Sim. & GPR & \% error \\
      \hline
      $F_p$ & 974.0 & 1017.6 & \textbf{4.43} & 971.0 & 1051.4 & \textbf{8.28} \\
      $\mathrm{CLE}$ & 0.562 & 0.579 & \textbf{3.11} & 0.569 & 0.578 & \textbf{1.74} \\
      $\mathrm{SEA}$ & 14.79 & 12.62 & \textbf{14.64} & 14.46 & 13.46 & \textbf{6.90} \\
      $\mathrm{\Delta Y_{node}}$ & 16.33 & 17.29 & \textbf{5.86} & 16.34 & 17.50 & \textbf{7.11} \\
      \hline
    \end{tabular}
    \caption{$n_{ls}$ = 4; $v_{p}$, $T_i$, $T_{pd}$, $T_{air}$  and $\mathbf{\Phi}_{fib}$ randomly selected (from table \ref{tab:doe_matrix}); $t_{l}$ different from training and testing }
    \label{tab:newdataset2}
\end{table}

\begin{table}[h]
    \centering
    \begin{tabular}{|c|c|c|c|c|c|c|}
     \hline
      Outputs & \multicolumn{3}{c|}{\shortstack{0.7 mm, 6.5 m/s,\\ 318 °C, 131 °C , 24 °C,\\ (30, -30, 60, -60)}} & \multicolumn{3}{c|}{\shortstack{0.9, 5.7 m/s,\\ 304 °C, 91 °C , 22 °C, \\ (30, -30, 60, -60)}}\\
      \hline
       & Sim. & GPR & \% error & Sim. & GPR & \% error \\
      \hline
      $F_p$ & 1020.0 & 1037.9 & \textbf{1.76} & 1010.0 & 1091.2 & \textbf{8.04} \\
      $\mathrm{CLE}$ & 0.540 & 0.577 & \textbf{6.9} & 0.554 & 0.582 & \textbf{4.98} \\
      $\mathrm{SEA}$ & 13.61 & 12.21 & \textbf{10.26} & 13.51 & 14.26 & \textbf{5.57} \\
      $\mathrm{\Delta Y_{node}}$ & 16.84 & 17.49 & \textbf{3.89} & 16.65 & 17.82 & \textbf{7.10} \\
      \hline
    \end{tabular}
    \caption{$n_{ls}$ = 4; $v_{p}$, $T_i$, $T_{pd}$, $T_{air}$  and $\mathbf{\Phi}_{fib}$ randomly selected (from range in table \ref{tab:doe_matrix}); $t_{l}$ different from training and testing}
    \label{tab:newdataset3}
\end{table}
\end{document}